\documentclass{article}

 \usepackage[main, final]{neurips_2026}
\usepackage{bm}
\usepackage[utf8]{inputenc} 
\usepackage[T1]{fontenc}    
\usepackage{hyperref}       
\usepackage{url}            
\usepackage{booktabs}       
\usepackage{amsfonts}       
\usepackage{nicefrac}       
\usepackage{float}
\usepackage{microtype}      
\usepackage{xcolor}         
\usepackage{graphicx}
\usepackage{amsmath} 
\usepackage{overpic}
\usepackage{xcolor}
\usepackage{colortbl}
\usepackage{subcaption}
\usepackage{wrapfig}
\usepackage{algorithm}
\usepackage{algpseudocode}
\algrenewcommand\algorithmicrequire{\textbf{Input:}}
\algrenewcommand\algorithmicensure{\textbf{Output:}}

\title{MaskSense: Confronting the Visual Exploration Trap \\ in  Masked Image Generation}

\author{Yawen Shao$^{1,2}$\thanks{~Work done at Tongyi Lab.} \quad Jie Xiao$^{2}$ \quad Kai Zhu$^{2}$\thanks{~Corresponding author.} \quad Yu Liu$^{2}$ \quad Hongchen Luo$^{3}$ \\ \textbf{Xueyang Fu}$^{1}$ \quad \textbf{Yang Cao}$^{1}$ \quad \textbf{Wei Zhai}$^{1}$\footnotemark[2] \quad \textbf{Zheng-Jun Zha}$^{1}$\\
{$^{1}$~University of Science and Technology of China} \\
{$^{2}$~Tongyi Lab, Alibaba Group} \qquad
{$^{3}$~Northeastern University} \\
}

\begin{document}

\maketitle

\begin{abstract}
Reinforcement learning (RL) has shown strong potential for aligning generative models with human intent, but adapting RL to masked generative models (MGMs) remains largely underexplored. 
Existing approaches that directly adapt Group Relative Policy Optimization (GRPO) to MGMs leave a foundational conflict unaddressed: MGMs' exploitative decoding nature is inherently at odds with the exploratory diversity that GRPO presupposes.
In this work, we reveal a structural decoding bias termed the \textbf{visual exploration trap}: confidence-based sampling trades the exploration of diverse subject realizations for the greedy resolution of low-uncertainty background regions, causing premature collapse of the generative exploration space and starving GRPO of the rollout diversity it depends on.
To this end, we propose \textbf{MaskSense}, a novel RL framework for MGMs that confronts the visual exploration trap at both the sampling and optimization stages. Specifically, we introduce a semantic anchored routing sampling mechanism that leverages semantic priors to preserve high-entropy subject token exploration while amplifying intra-group reward variance through exploratory-exploitative routing 
to yield more discriminative advantage estimates.  
Furthermore, we design a global entropy transition anchoring strategy to identify the most consequential decoding steps and concentrate gradient updates on them. 
Extensive experiments on multiple text-to-image benchmarks demonstrate that MaskSense substantially improves upon the base model, with GenEval accuracy improving from 54\% to 86\% and HPS from 28.89 to 36.39, achieving state-of-the-art performance. 


%
%
%
\end{abstract}   
\section{Introduction}
\label{sec:intro}

Masked generative models (MGMs)~\cite{bai2024meissonic,chang2023muse,luminadimoo,maskedlearners} have recently emerged as a compelling paradigm for visual synthesis, challenging the long-standing dominance of autoregressive models (AR)~\cite{wu2025janus,liu2024lumina,llama}  and diffusion models (DMs)~\cite{ddim,ddpm,lipman2022flow} by reformulating image generation as iterative masked token prediction.
%
Central to this paradigm is the confidence-based sampling strategy of MaskGIT~\cite{chang2022maskgit}, which predicts all masked tokens in parallel at each step and retains only the most confident subset while remasking the remainder for subsequent iterations, making the decoding process exploitative.
 
Reinforcement learning (RL)~\cite{grpo,ppo,dpo} has proven 
highly effective in enhancing the reasoning capabilities of 
large language models (LLMs)~\cite{guo2025deepseek,openai,20/80,shao2026trackaligningrewardsstates}, while its application to MGMs 
remains comparatively underexplored relative to diffusion 
models~\cite{flow-grpo,vgpo,xue2025dancegrpo} and AR models~\cite{T2I-R1,gcpo}.
Recent works~\cite{zhang2025maskfocus,mask-grpo} have begun 
addressing this gap by applying Group Relative Policy 
Optimization (GRPO) to MGMs, yet retain the confidence-based 
sampler as a fixed component with minimal adaptation to the 
masked generation setting.
This overlooks a fundamental prerequisite of GRPO: its 
effectiveness derives from group-relative advantage estimation, 
which presupposes rollouts sufficiently diverse to cover 
meaningful regions of the solution space, a requirement 
inherently at odds with the exploitative nature of 
confidence-based sampling.

\begin{figure*}[t]
\centering
    \begin{overpic}[width=1\linewidth]{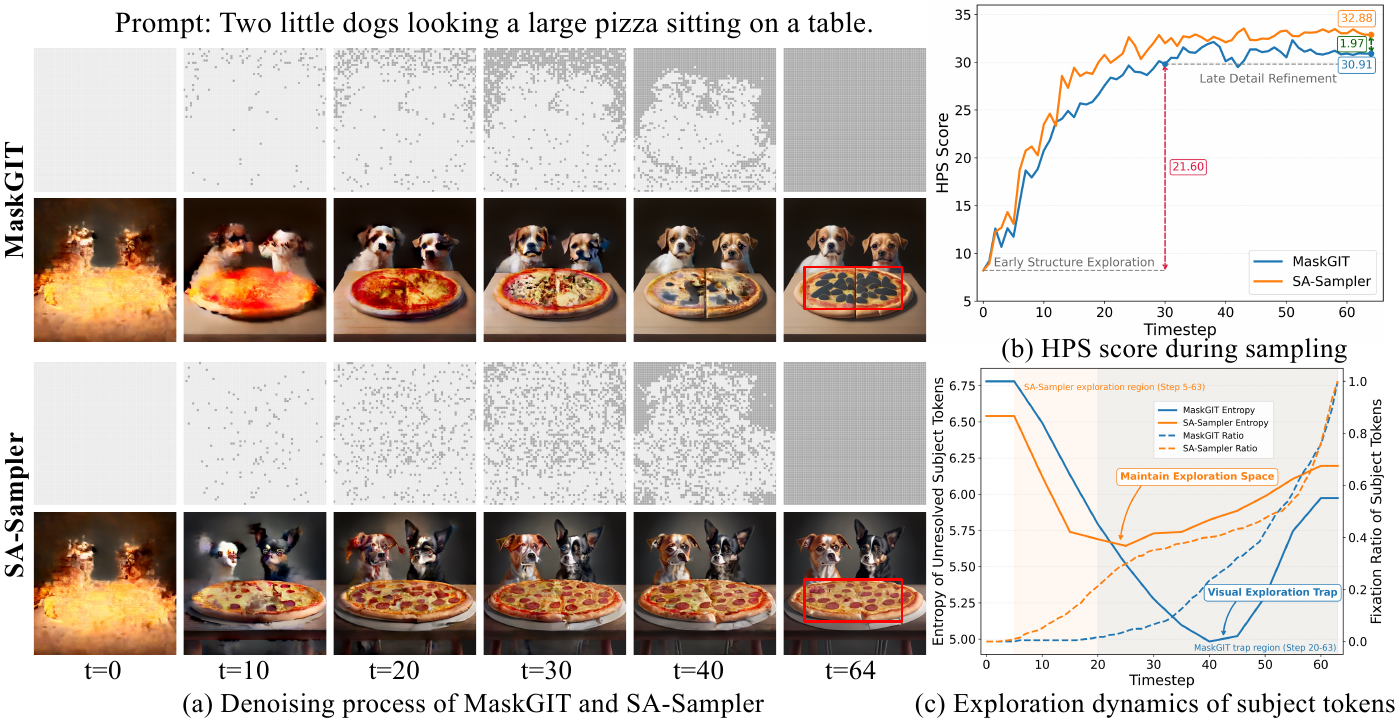}  
    \end{overpic}
    
    \caption{\small\textbf{Motivation.} (a) MaskGIT keeps subject tokens masked while committing background regions early, whereas SA-Sampler resolves subject regions earlier via semantic anchoring. (b) SA-Sampler consistently discovers higher-reward trajectories through broader generative exploration. (c) SA-Sampler maintains higher subject-token fixation ratios and elevated entropy, effectively escaping the visual exploration trap.}
    \label{fig:motivation}
\vspace{-3.5mm}
\end{figure*}

In this paper, we identify and characterize precisely how 
this conflict manifests as a structural bias that we term 
the \textbf{visual exploration trap}. Specifically, tokens with lower uncertainty predominantly correspond to simple background textures in natural images~\cite{ma2025towards}, causing background regions to be committed well before semantically critical subject areas are ever explored (Fig.~\ref{fig:motivation} (a)).
%
Subject tokens, by contrast, retain high predictive entropy throughout the early generation steps, while their fixation ratio remains near zero (Fig.~\ref{fig:motivation} (c)), indicating that substantial generative diversity is available yet remains entirely unexploited.
By the time subject tokens begin to be retained after step 20, 
the surrounding background context has been largely 
committed, severely narrowing the space of potential generative branches~\cite{justgrpo,20/80} and limiting the diversity and quality of the resulting subject regions~\cite{ma2025towards}.
The originally high entropy of subject tokens is thus suppressed before any meaningful subject decision is made, as the model effectively sacrifices diverse subject realizations to greedily resolve low-uncertainty background regions, directly undermining the rollout diversity that GRPO requires.  This trap necessitates a rethink of RL for MGMs.
%
%

To tackle this challenge, we propose \textbf{MaskSense} (Fig.~\ref{fig:method}), a novel RL framework designed for MGMs that directly confronts the visual exploration trap at both the sampling and optimization stages.
At the sampling stage, we propose the semantic anchored routing sampling (SARS) mechanism, built upon the Semantic Anchored Sampler (SA-Sampler), which employs semantic priors to anchor subject tokens during the early generation steps (Fig.~\ref{fig:motivation} (a)). This design preserves the high-entropy exploration space that confidence-based sampling would otherwise suppress (Fig.~\ref{fig:motivation} (c)), enabling the discovery of higher-reward trajectories 
through broader generative exploration (Fig.~\ref{fig:motivation} (b)).
SARS further routes rollouts stochastically between SA-Sampler and MaskGIT, generating a mixture of exploratory and exploitative trajectories within each group to deliberately amplify intra-group reward variance and sharpen discriminative advantage estimates. 
At the optimization stage, we further introduce global 
entropy transition anchoring (GETA) to concentrate gradient 
updates on the most structurally decisive decoding steps, 
avoiding diffuse optimization across steps of unequal 
importance.
%
%
%
%
Extensive experiments on multiple text-to-image benchmarks validate the effectiveness of MaskSense.

The contributions of this paper are summarized as follows:
\begin{itemize}
\item We identify the visual exploration trap in MGMs, revealing that confidence-based sampling suppresses subject-token entropy before meaningful subject decisions are made, fundamentally  depriving GRPO of the exploratory diversity essential to its optimization.
\item We propose MaskSense, a novel RL framework for MGMs that promotes diverse rollout exploration through semantic anchored routing sampling, and concentrates gradient updates on the most consequential decoding timesteps via global entropy transition anchoring.
\item Extensive experiments on multiple text-to-image benchmarks demonstrate that MaskSense achieves state-of-the-art performance, 
confirming the effectiveness of confronting the visual exploration trap for RL optimization in MGMs.
\end{itemize}
  
\section{Related Work}
\label{sec:related_work}

\textbf{Visual Generation Models.} The rapid advancement of visual generation~\cite{wan2025wan,flux,bai2024meissonic,
wu2025janus,show-o, PEBench, zhu2026doesvisualgenerationhelp, song2026stylevideo,xiao2026divideconquerdecoupledrepresentation,wu2024improvedvideovaelatent,wan2026} has been propelled by three dominant architectures: DMs, AR, and MGMs. 
DMs~\cite{ddpm,ddim,ldm,sdxl,lipman2022flow,lu2025dpm} reframe generation as an iterative denoising process, which suffers from slow inference. AR ~\cite{emu,Janus-Pro,Image-Transformer,Esser_2021_CVPR, wu2026sequencemodelingalignmenttokenizer, Dai_2025, yang2025mmarlosslessmultimodalautoregressive} predict discrete tokens sequentially via causal transformers, but their raster-order nature incurs prohibitive computational overhead at high resolution~\cite{NEURIPS2019_5f8e2fa1}. To address the limitations of both paradigms, MGMs~\cite{chang2023muse,chang2022maskgit,luminadimoo,bai2024meissonic} cast image synthesis as an iterative mask prediction problem, where all tokens are decoded in parallel at each step and progressively refined conditioned on previously unmasked tokens. MaskGIT~\cite{chang2022maskgit} pioneered this direction by introducing bidirectional masked modeling with effective mask scheduling and parallel decoding, establishing the foundational framework for subsequent works. Meissonic~\cite{bai2024meissonic} further advances this paradigm through hybrid transformer architectures and feature compression, successfully scaling masked generative modeling to high-resolution text-to-image generation. The efficiency and growing capability of MGMs motivate further exploration of their potential through RL-based optimization.

\textbf{Reinforcement Learning for Text-to-Image Generation.}
With recent advances in large language and multimodal modeling~\cite{zhu-etal-2025-llm,zhu2026wanpecinematicpromptenhancement,kan2026vivasvitalizingvisualperception,kan2026uvuimprovingmultimodalunderstanding,wei2026youtuvlunleashingvisualpotential,lu2025benchmarkinglargevisionlanguagemodels,shao2025greatgeometryintentioncollaborativeinference,sun2024moss}, reinforcement learning (RL)~\cite{grpo,ppo,dpo} has proven effective in enhancing the reasoning capabilities of these models~\cite{openai,zhao2025d1scalingreasoningdiffusion,cui2025processreinforcementimplicitrewards,kan2025tacothinkanswerconsistencyoptimized}, spurring its rapid adoption in text-to-image generation.
Prior works primarily target preference alignment in diffusion-based models~\cite{d-dpo,ddpo}, while more recent efforts successfully extend GRPO-based optimization to flow matching~\cite{flow-grpo,xue2025dancegrpo,vgpo} and AR models~\cite{T2I-R1,gcpo}. Within MGMs, Mask-GRPO~\citep{mask-grpo} formulates the iterative unmasking process as a multi-step decision-making problem and proposes transition probability to enable GRPO optimization. 
MaskGRPO~\citep{maskgrpo} introduces modality-aware importance estimation to improve the image generation quality of multimodal discrete diffusion models. MaskFocus~\citep{zhang2025maskfocus} identifies sampling steps via inter-step embedding similarity and incorporates dynamic sampling. However, none of these methods escape the confidence-based sampling strategy, 
leaving the visual exploration trap and principled gradient allocation across decoding steps fundamentally unaddressed as core challenges in RL optimization for MGMs. 
\section{Preliminaries}
\label{sec:preliminaries}
\subsection{Masked Generative Models}
Masked generative models (MGMs) can be formulated as discrete diffusion processes, in which a clean token sequence $x_0$ is progressively corrupted into either a new token or an absorbed state $x_t$ at step $t$ via a Markov process, and a learned reverse process subsequently reconstructs the original data from these corrupted observations.
Specifically, the corruption mechanism operates over discrete tokens through a transition matrix ${Q}_t$, constructed as a sequential product of per-step matrices ${Q}_t = \prod_{s=1}^{t} {Q}_{s}$. The forward process is formulated as:
\begin{equation}
\label{eq1}
x_t \sim q(x_t|x_0,t),\quad
q(x_t|x_0,t)=\text{Cat}(x_t;\, x_0{Q}_t)
\end{equation}
where ${Q}_t$ encodes structured transition patterns, such as Gaussian kernels, nearest-neighbor mappings, or absorbing states. Meissonic adopts the absorbing state formulation, in which each token is independently replaced with a \texttt{[MASK]} token and remains in the absorbed state once masked.

To recover the original sequence, the reverse process $p_\theta(x_{t-1}|x_t)$ is modeled by a network that learns to progressively denoise the corrupted sequence, starting from the fully masked state $x_T$ and iteratively transitioning toward the clean data $x_0$. The reverse process is formulated as:
\begin{equation}
\label{eq2}
    p_\theta(x_{t-1}| x_t) = \sum_{x_0} q(x_{t-1}| x_t, x_0)\, p_\theta(x_0| x_t)
\end{equation}
During inference, generation begins from a fully \texttt{[MASK]} canvas and proceeds over $T$ iterative decoding steps~\cite{bai2024meissonic}. At each step, the model predicts all masked tokens in parallel, commits a high-confidence subset, and remasks the remaining positions for subsequent refinement.

\subsection{Group Relative Policy Optimization}
Group Relative Policy Optimization (GRPO)~\cite{grpo} is a reinforcement learning algorithm that leverages the average reward of multiple sampled outputs as a dynamic baseline for advantage estimation. This principle has been recently adapted for masked generative models in Mask-GRPO~\cite{mask-grpo}, which applies GRPO to improve the performance of state-of-the-art MGMs~\cite{bai2024meissonic,show-o}. 
Given a prompt $c$, the policy model $p_{\theta}$ samples a group of $G$ images $\left\{x_{0}^{i}\right\}_{i=1}^{G}$ along with the corresponding reverse-time trajectories $\left\{x_{T}^{i},x_{T-1}^{i},\cdots,x_{0}^{i}\right\}_{i=1}^{G}$. The advantage of the $i$-th image is then calculated by normalizing the group-level rewards as follows:
\begin{equation}
\label{eq3}
\hat{A}_{t}^{i}=\frac{R\left(x_{0}^{i}, c\right)-\operatorname{mean}\left(\left\{R\left(x_{0}^{i}, c\right)\right\}_{i=1}^{G}\right)}{\operatorname{std}\left(\left\{R\left(x_{0}^{i}, c\right)\right\}_{i=1}^{G}\right)}
\end{equation}

Mask-GRPO~\cite{mask-grpo} optimizes the policy model by maximizing the following objective:
\begin{equation}
\label{eq4}
\begin{array}{l}
\mathcal{J}_{\text {GRPO}}(\theta)=\mathbb{E}_{c \sim \mathcal{C},\left\{x^{i}\right\}_{i=1}^{G} \sim \pi_{\theta_{\text {old }}}(\cdot \mid c)} \\
\qquad\left[\frac{1}{G} \frac{1}{T} \sum_{i=1}^{G} \sum_{t=1}^{T}\left(\min \left(r_{t}^{i}(\theta) \hat{A}_{t}^{i}, \operatorname{clip}\left(r_{t}^{i}(\theta), 1-\varepsilon, 1+\varepsilon\right) \hat{A}_{t}^{i}\right)-\beta D_{\mathrm{KL}}\left(\pi_{\theta} \| \pi_{\mathrm{ref}}\right)\right)\right]
\end{array}
\end{equation}
where the probability ratio $r_{t}^{i}(\theta)$ is defined as:
\begin{equation}
\label{eq5}
r_{t}^{i}(\theta)=\frac{p_{\theta}\left(x_{t-1}^{i} \mid x_{t}^{i}, c\right)}{p_{\theta_{\text {old }}}\left(x_{t-1}^{i} \mid x_{t}^{i}, c\right)}
\end{equation}
  
\section{Method}
\label{sec:method}

\subsection{Overview}
\label{sec:method_overview}

\begin{figure*}[t]
\centering
\begin{overpic}[width=1\linewidth]{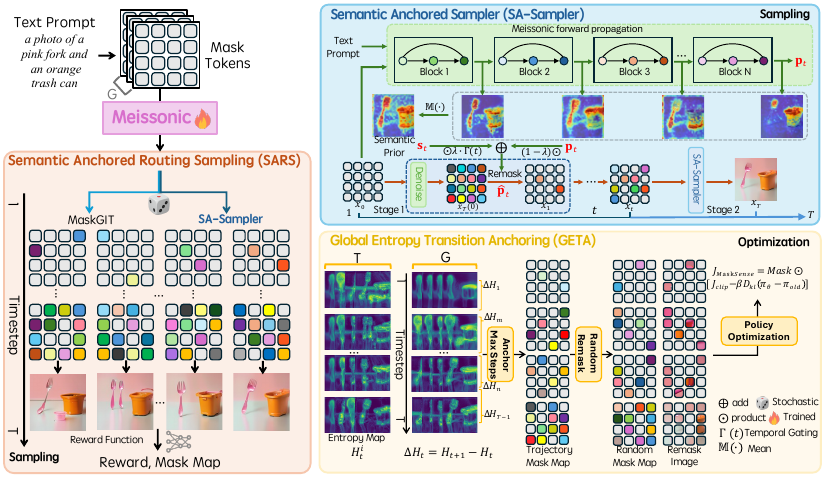}
\end{overpic}
\caption{\textbf{Overview of MaskSense.} MaskSense comprises two 
core modules that jointly address the visual exploration trap. 
At the sampling stage, SARS promotes subject-token exploration via semantic anchored decoding and sharpens advantage estimation through dual-branch rollout routing. At the optimization stage, GETA concentrates GRPO updates on critical decoding steps anchored by entropy transitions, with random remasking applied to ensure optimization diversity.}
\label{fig:method}
\end{figure*}

MaskSense addresses a fundamental mismatch in applying GRPO to masked generative models: GRPO requires diverse rollout groups, whereas confidence-based decoding in MaskGIT prematurely commits to low-uncertainty trajectories, suppressing exploration. Meanwhile, applying GRPO uniformly across all decoding steps ignores their temporally uneven contribution to generation quality.
 MaskSense resolves this tension via two core modules, as shown in Fig.~\ref{fig:method}. \textbf{Semantic Anchored Routing Sampling} (SARS) introduces a routing strategy that integrates a novel semantic anchored sampler (SA-Sampler) alongside MaskGIT to balance exploration and exploitation, while \textbf{Global Entropy Transition Anchoring} (GETA) identifies the decoding timesteps with the largest entropy transition as optimization anchors, focusing GRPO updates where token commitment is most decisive.

\subsection{Semantic Anchored Routing Sampling}
\label{sec:sars}

SARS targets the visual exploration trap, in which confidence-driven decoding preferentially resolves low-uncertainty background regions while deferring commitment on semantically decisive subject tokens. To counteract this bias, SARS derives a prompt-conditioned semantic prior, injects it into the early token retention process, and routes rollout groups through a dual-branch structure.

\textbf{Semantic Prior Construction.}
We construct the semantic prior from image-to-text cross-attention cached at decoding step $t$, where $t \in \{0, 1, \dots, T\}$ denotes the number of completed decoding steps in contrast to $t$ in Sec.~\ref{sec:preliminaries}. Let $\mathbf{F} \in \mathbb{R}^{B \times N_{\mathrm{img}} \times N_{\mathrm{txt}}}$ denote the cross-attention map extracted from each transformer block, where $B$ is the batch size, $N_{\mathrm{img}}$ and $N_{\mathrm{txt}}$ denote the numbers of image and text tokens, respectively.
We average over image positions to compute a global response score $\Phi_j$ for each text token $j \in \{1, \dots, N_{\mathrm{txt}}\}$,
and select the top-$k$ tokens with the highest scores as semantic anchors
$\mathcal{T} \subset \{1,\dots,N_{\mathrm{txt}}\}$, $|\mathcal{T}|=k$. The response score and importance weight are formulated as:
\begin{equation}
    \Phi_j = \frac{1}{N_{\mathrm{img}}} \sum_{u=1}^{N_{\mathrm{img}}} \mathbf{F}_{u,j},
    \quad
    w_n = \frac{\Phi_n}{\sum_{n \in \mathcal{T}} \Phi_n}
\end{equation}
where $\Phi_j$ quantifies the degree to which text token $j$ is attended to across all image positions, and $w_n$ denotes the normalized importance weight of anchor token $n$.
The block-level semantic map $\mathbf{s}_b \in \mathbb{R}^{B \times N_{\mathrm{img}}}$ and the final semantic prior $\mathbf{s}_t \in [0,1]^{B \times N_{\mathrm{img}}}$ are subsequently computed as:
\begin{equation}
\label{eq7}
    \mathbf{s}_b = \sum_{n \in \mathcal{T}} w_n\, \mathbf{F}_{n},
    \quad
    \mathbf{s}_t = \Psi\!\left(\frac{1}{N_b}\sum_{b=1}^{N_b} \mathbf{s}_b\right)
\end{equation}
where $\mathbf{F}_{n} \in \mathbb{R}^{B \times N_{\mathrm{img}}}$ denotes the attention slice of text token $n$ along the text token dimension of $\mathbf{F}$, $N_b$ is the number of transformer blocks, and $\Psi(\cdot)$ denotes percentile normalization onto $[0,1]$. 

\textbf{Semantic Anchored Sampler (SA-Sampler).}
Since early steps determine the global structure of the generated image while late-stage injection risks disrupting fine-grained detail refinement (Fig.~\ref{fig:motivation} (b)), we confine semantic prior guidance to the early decoding stage via a linearly decaying temporal gate:
\begin{equation}
\label{eq8}
    \Gamma(t) = \max\!\left(0,\, 1 - \frac{r_t}{\tau}\right), 
    \quad r_t = \frac{t}{T}
\end{equation}
where $r_t$ denotes the decoding progress ratio, $T$ is the total number of decoding steps, and $\tau \in (0,1]$ defines the boundary of the early decoding stage within which semantic anchoring is active.

At each step $t$, the model predicts token probabilities $\mathbf{p}_t$ as confidence scores for all masked positions in parallel, and attains the projected terminal state $\hat{x}_T(t)$. Rather than modifying the predictor, we reweight this score with the semantic prior to obtain the semantic anchored retention score $\hat{\mathbf{p}}_t$:
\begin{equation}
\label{eq9}
    \hat{\mathbf{p}}_t = (1-\lambda)\,\mathbf{p}_t + \lambda\,\Gamma(t)\,\mathbf{s}_t
\end{equation}
where $\lambda$ controls the strength of semantic guidance. The reweighted score $\hat{\mathbf{p}}_t$ is used exclusively to determine which tokens are retained as $x_{t+1}$ and which are remasked.

\textbf{Dual-Branch Routing.} To balance exploration and exploitation within each GRPO rollout group, we stochastically partition 
the $G$ sampled trajectories for each prompt into a semantic branch $\mathcal{G}_{\mathrm{sa}}$ (SA-Sampler) and a standard branch 
$\mathcal{G}_{\mathrm{mg}}$ (MaskGIT sampler), with $|\mathcal{G}_{\mathrm{sa}}| = |\mathcal{G}_{\mathrm{mg}}| = \frac{G}{2}$ 
and branch assignments fixed throughout the entire decoding process.
By coupling exploratory and conservative trajectories within the same group, this design diversifies sampling outcomes, amplifies intra-group reward contrast, and yields more discriminative advantage estimates for GRPO.

\subsection{Global Entropy Transition Anchoring}
\label{sec:geta}
Even with improved rollout diversity, denoising steps contribute unevenly to policy learning: some involve only local token refinements, while others drive sharp collective uncertainty shifts that fundamentally reshape the generation trajectory, manifesting as large entropy transitions across denoising steps. GETA exploits this signal to concentrate GRPO updates on critical steps.

\textbf{Entropy Transition Anchor Selection.}
GETA measures the magnitude of change in the model's global uncertainty across decoding steps. For each rollout, we record the token distribution at every timestep $t$ and compute the group-averaged entropy, formulated as:
\begin{equation}
\label{eq10}
    {H}_t = -\frac{1}{G}\sum_{i=1}^{G}
    \sum_{v\in\mathcal{V}} p^i_t(v)\log p^i_t(v)
\end{equation}
where $\mathcal{V}$ is the codebook and $p^i_t(v)$ is the predicted probability of token $v$ at step $t$ in the $i$-th rollout. 

The entropy transition magnitude between consecutive steps is then computed, and the top-$k$ steps with the largest transitions are selected as anchor steps:
\begin{equation}
\label{eq11}
    \Delta H_t = \left|{H}_{t+1}-{H}_t\right|, \quad
    \mathcal{K}=\mathrm{topk}\big(\{\Delta H_t\}_{t=1}^{T-1}, k\big)
\end{equation}
where $\mathcal{K}$ denotes the set of selected anchor steps and $k$ is the number of anchor steps to select.

\textbf{Random Remasking for Optimization.} Once the anchor set $\mathcal{K}$ is obtained, we construct optimization states via random remasking~\cite{zhang2025maskfocus}. 
Let $M_t^{\mathrm{traj}}$ denote the binary mask at anchor step $t$ in the original rollout. Rather than replaying this, we sample a random mask $M_t^{\mathrm{rand}}$ with the same mask budget:
\begin{equation}
\label{eq12}
\left|M_t^{\mathrm{rand}}\right| = \left|M_t^{\mathrm{traj}}\right|,\quad
M_t^{\mathrm{rand}} \sim \mathcal{U}\Big(\{M\in\{0,1\}^{H\times W}:\ |M|=\left|M_t^{\mathrm{traj}}\right|\}\Big),
\end{equation}
where $\mathcal{U}(\cdot)$ denotes the uniform distribution, $H\times W$ is the spatial resolution of the latent token map.
This random mask is then applied to the final generated latent $x_T$ to produce the remasked input:
\begin{equation}
\label{eq13}
\tilde{x}_t = M_t^{\mathrm{rand}}\odot [\texttt{MASK}] + 
\left(1-M_t^{\mathrm{rand}}\right)\odot x_T,
\end{equation}
where $\odot$ denotes element-wise product. By decoupling 
optimization from the exact trajectory mask pattern while preserving the number of masked positions, the policy is trained around each critical transition without overfitting to any single realized masking configuration.

The remasked input $\tilde{x}_t$ is subsequently used to compute the importance ratio $r_t^i(\theta)$ in Eq.~\ref{eq5}, and the GRPO objective (Eq.~\ref{eq4}) is optimized exclusively over the selected anchor steps $t \in \mathcal{K}$:
\begin{equation}
\label{eq14}
\mathcal{J}_{\text{MaskSense}}(\theta) = M_t^{\mathrm{rand}} \odot\, 
\mathcal{J}_{\text{GRPO}}(\theta)\big|_{t \in \mathcal{K}},
\end{equation}
where the mask $M_t^{\mathrm{rand}}$ restricts gradient updates to the masked positions of the remasked input $\tilde{x}_t$, ensuring that only the tokens requiring prediction contribute to the optimization objective. The complete MaskSense algorithm is detailed in Algorithm \ref{alg:masksense} in appendix.

\section{Experiments}
\definecolor{mygray}{gray}{.92}
\definecolor{deltagreen}{HTML}{228B22}

\newcommand{\inc}[1]{\textcolor{deltagreen}{+#1}}

\begin{table}[t]
    \caption{
        \textbf{Quantitative Comparison on the GenEval.} Best scores are \textbf{bolded}. Results for other models are from~\cite{zhang2025maskfocus}. $\Delta$ indicates improvements over Meissonic. Obj.: Object; Attr.: Attribution.
    }
    \label{tab:geneval}
    \centering
    \footnotesize
    \renewcommand{\arraystretch}{1}
    \renewcommand{\tabcolsep}{6pt}
    \resizebox{\linewidth}{!}{
    \begin{tabular}{lccccccc}
        \toprule
        \textbf{Model} & \textbf{Overall}$\uparrow$ & \textbf{Single Obj.}$\uparrow$ & \textbf{Two Obj.}$\uparrow$ & \textbf{Counting}$\uparrow$ & \textbf{Color}$\uparrow$ & \textbf{Position}$\uparrow$ & \textbf{Color Attr.}$\uparrow$ \\
        \midrule
        \multicolumn{8}{c}{\textit{Diffusion-based Models}} \\ \midrule 
        LDM~\cite{ldm}                 & 0.37 & 0.92 & 0.29 & 0.23 & 0.70 & 0.02 & 0.05 \\
        SD-XL~\cite{sdxl}              & 0.55 & 0.98 & 0.74 & 0.39 & 0.85 & 0.15 & 0.23 \\
        SD3.5-M~\cite{sd3.5}           & 0.63 & 0.98 & 0.78 & 0.50 & 0.81 & 0.24 & 0.52 \\
        DALLE-3~\cite{dalle3}          & 0.67 & 0.96 & 0.87 & 0.47 & 0.83 & 0.43 & 0.45 \\ 
        FLUX.1-dev~\cite{flux}         & 0.66 & 0.98 & 0.81 & 0.74 & 0.79 & 0.22 & 0.45 \\
        \midrule
        
        \multicolumn{8}{c}{\textit{AR-based Models}} \\ \midrule
        LlamaGen~\cite{llama}          & 0.32 & 0.71 & 0.34 & 0.21 & 0.58 & 0.07 & 0.04 \\
        JanusFlow~\cite{ma2025janusflow}& 0.63 & 0.97 & 0.59 & 0.45 & 0.83 & 0.53 & 0.42 \\
        Janus-Pro-1B~\cite{Janus-Pro}  & 0.73 & 0.98 & 0.82 & 0.51 & \textbf{0.89} & 0.65 & 0.56 \\
        Emu3($+$Rewriter)~\cite{emu3}  & 0.66 & 0.99 & 0.81 & 0.42 & 0.80 & 0.49 & 0.45 \\
        \midrule

        \multicolumn{8}{c}{\textit{Masked Generative Models}} \\ \midrule
        Show-o~\cite{show-o}                         & 0.68 & 0.98 & 0.80 & 0.66 & 0.84 & 0.31 & 0.50 \\
        Show-o$+$PARM~\cite{showparm}                & 0.69 & 0.97 & 0.75 & 0.60 & 0.83 & 0.54 & 0.53 \\
        Mask-GRPO~\cite{mask-grpo}                   & 0.73 & 0.99 & 0.90 & 0.69 & 0.85 & 0.35 & 0.59 \\
        Meissonic~\cite{bai2024meissonic}            & 0.54 & 0.99 & 0.66 & 0.42 & 0.86 & 0.10 & 0.22 \\
        \quad$+$ GRPO~\cite{grpo}                    & 0.72 & 0.98 & 0.84 & 0.76 & 0.85 & 0.45 & 0.41 \\
        \quad$+$ MaskGRPO~\cite{maskgrpo}            & 0.73 & \textbf{1.00} & 0.87 & 0.83 & 0.87 & 0.39 & 0.48 \\
        \quad$+$ MaskFocus~\cite{zhang2025maskfocus} & 0.76 & \textbf{1.00} & 0.91 & 0.85 & 0.87 & 0.42 & 0.54 \\
        \rowcolor{mygray}
        \textbf{\quad$+$ MaskSense (Ours)}
            & \textbf{0.86}
            & 0.99
            & \textbf{0.93}
            & \textbf{0.86}
            & \textbf{0.89}
            & \textbf{0.86}
            & \textbf{0.65} \\
        \rowcolor{mygray}
        \textcolor{deltagreen}{\quad$\Delta$ Over Base Model}
            & \inc{0.32}
            & \inc{0.00}
            & \inc{0.27}
            & \inc{0.44}
            & \inc{0.03}
            & \inc{0.76}
            & \inc{0.43} \\
        \bottomrule
    \end{tabular}
    }
\end{table}

\subsection{Experimental Setup} 
\textbf{Training Settings.} All experiments are conducted on  Meissonic~\cite{bai2024meissonic}, a masked text-to-image model at $1024 \times 1024$ resolution, trained on 8 NVIDIA A100 GPUs. Following the standard masked-generation RL setup~\cite{zhang2025maskfocus}, we adopt a classifier-free guidance scale of $5$ during both training and inference, perform 64 iterative decoding steps, and set the number of selected optimization anchors to 6. For the compositional image generation task~\cite{ghosh2023geneval}, we use 50K GenEval-style prompts sourced from Flow-GRPO~\cite{flow-grpo}. For preference alignment, we use 10K training samples from HPS v2 dataset~\cite{hps}. 
More implementation details are provided in Appendix~\ref{A}.

\textbf{Evaluation Details.} We evaluate MaskSense from two complementary perspectives: (i) \textit{task-specific accuracy} on in-distribution test sets, which directly measures the model's ability to handle the target task. (ii) \textit{general image quality} on DrawBench~\cite{drawbench}, which assesses broader generalization beyond the training distribution. The latter is quantified using a comprehensive set of metrics, including image quality (DeQA~\cite{deqa}) and human preference scores (PickScore~\cite{pickscore}, ImageReward~\cite{imagereward}).

\begin{figure}[t]
  \centering

  \begin{subfigure}{0.32\linewidth}
  \centering
    \includegraphics[width=\linewidth]{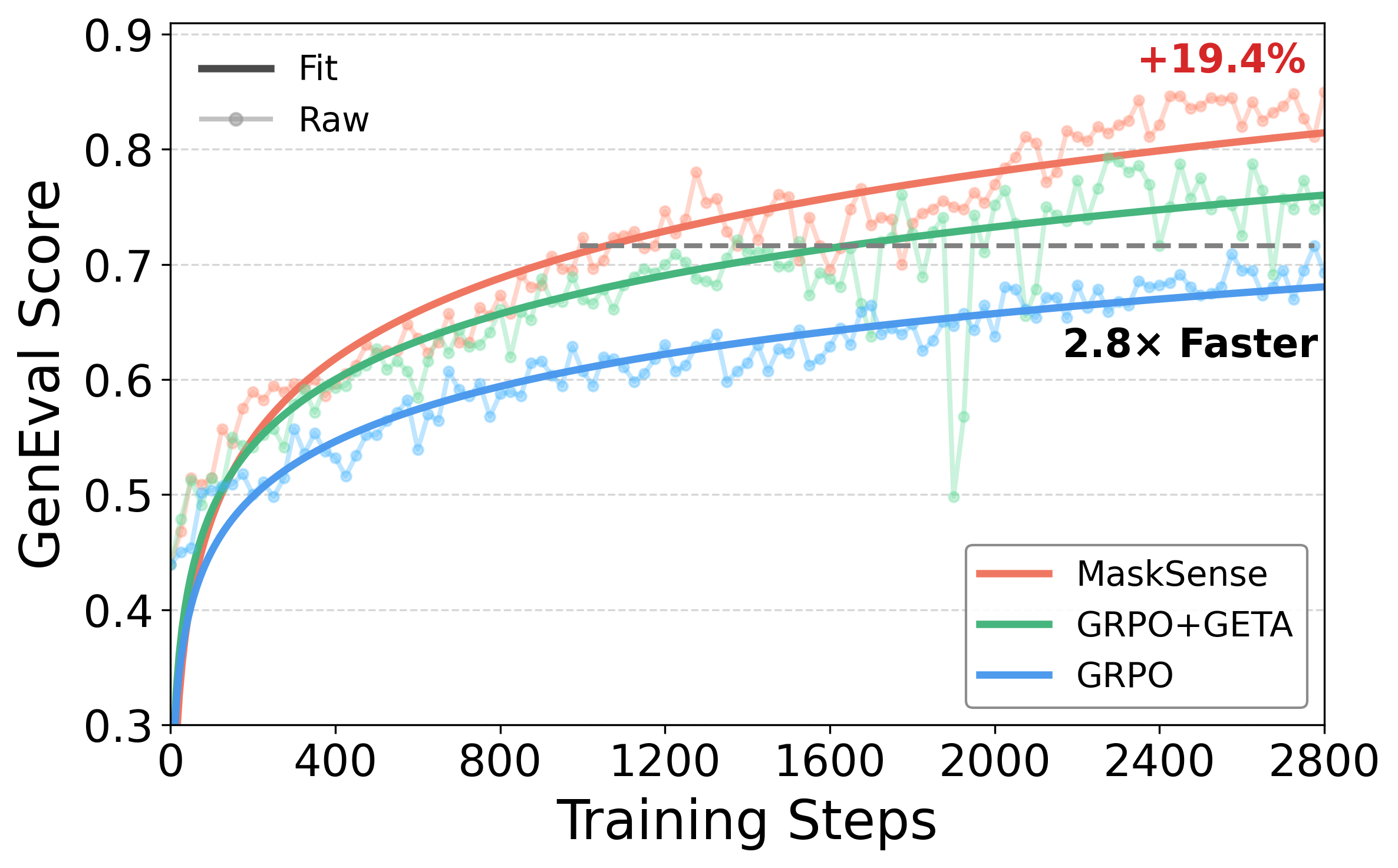}
    \caption{Compositional Image Gen.}
    \label{fig:geneval}
  \end{subfigure}
  \hfill
  \begin{subfigure}{0.32\linewidth}
  \centering
    \includegraphics[width=\linewidth]{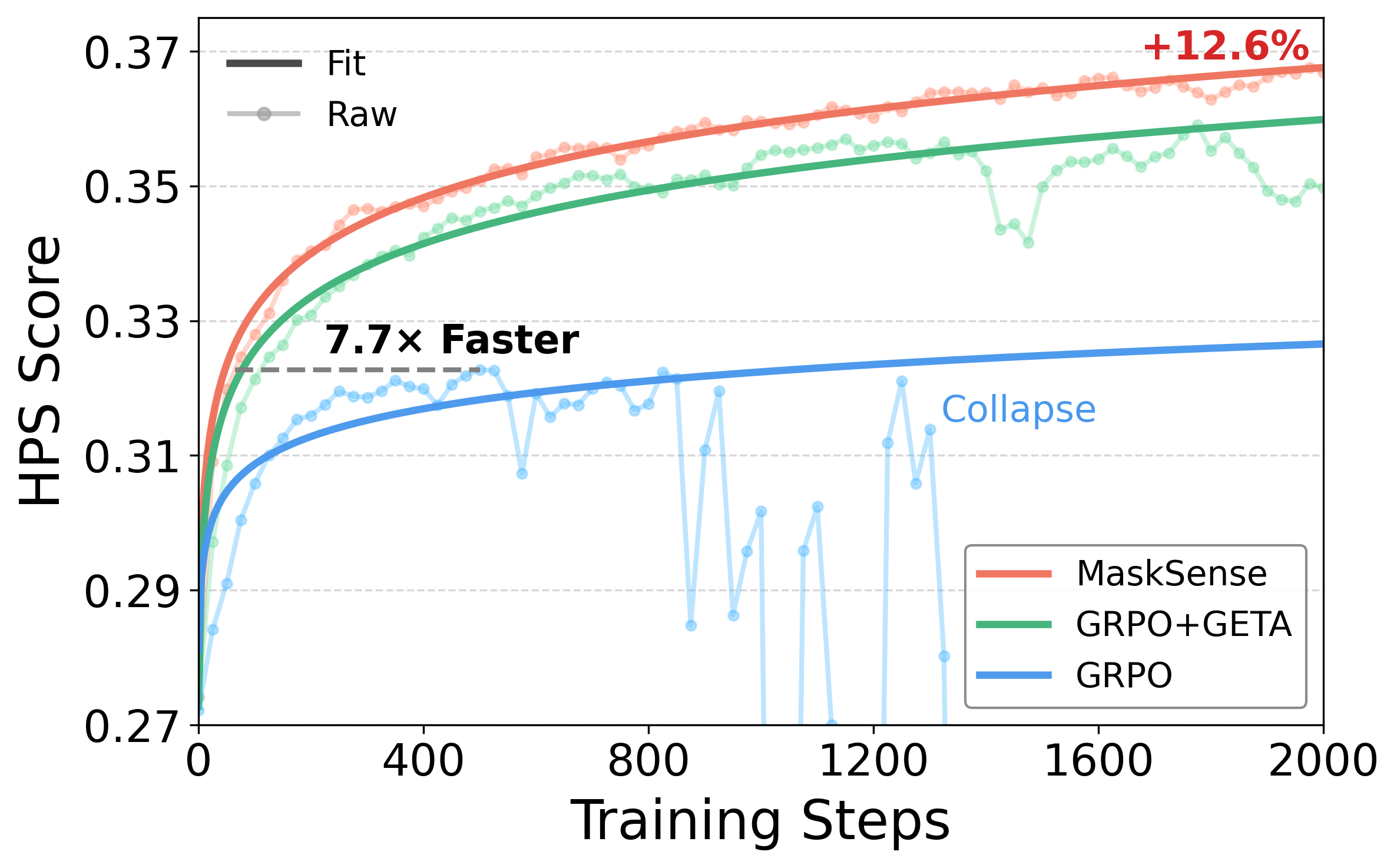}
    \caption{Human Preference Alignment}
    \label{fig:hps}
  \end{subfigure}
  \hfill
  \begin{subfigure}{0.32\linewidth}
  \centering
    \includegraphics[width=\linewidth]{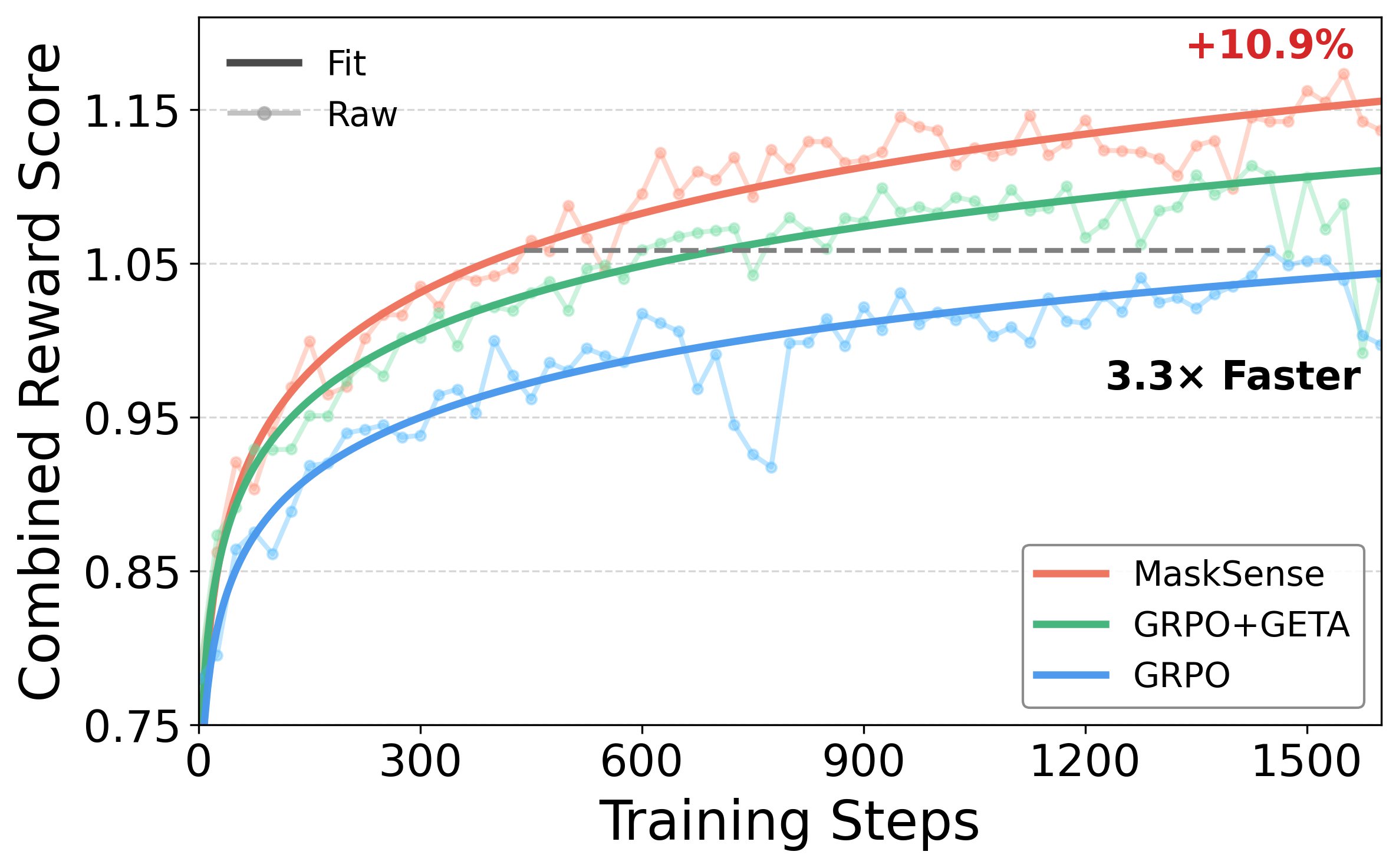}
    \caption{Multi-Reward Optimization}
    \label{fig:multi-reward}
  \end{subfigure}
  \caption{\textbf{Comparison of Training Dynamics Curves.} Fig.(a) to (c) refer to the tasks of compositional image generation, human preference alignment, and multi-reward optimization, respectively.}
  \label{fig:training}

\end{figure}

\subsection{Main Results} 
\textbf{Quantitative Comparison.}
Tab.~\ref{tab:geneval} and ~\ref{table:hps} demonstrate that MaskSense achieves superior performance across both tasks, improving GenEval by 0.32 and HPS by 7.50 over the Meissonic baseline.
%
Furthermore, these gains generalize beyond the training reward domain, as MaskSense achieves the best scores across DrawBench metrics, confirming broader generalization. 
As shown in Fig.~\ref{fig:training}, MaskSense exhibits superior training stability and faster convergence to a higher final performance.

\begin{figure}[t]
\centering
\begin{overpic}[width=1\linewidth]{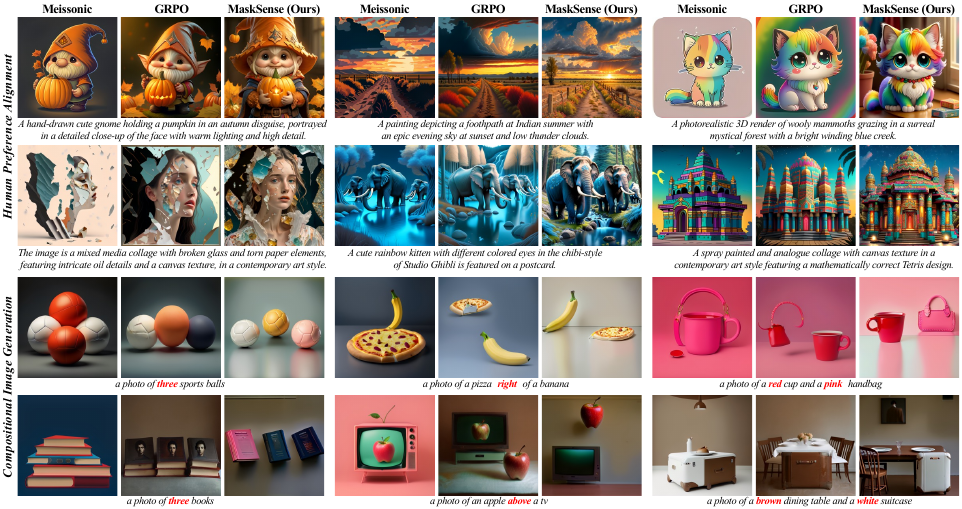}
\end{overpic}
\caption{\textbf{Qualitative Comparison.} MaskSense achieves superior visual quality with richer fine-grained details and more faithful satisfaction of object-level constraints across both tasks.}
\label{fig:result}
\end{figure}

\textbf{Qualitative Comparison.}
Fig.~\ref{fig:result} provides qualitative evidence consistent with the quantitative results. MaskSense produces richer fine-grained details, complex textures, and higher visual fidelity for preference-oriented prompts, while other methods yield flatter appearance and incomplete stylistic rendering. For compositional prompts, MaskSense more reliably satisfies object-level constraints including count, spatial arrangement, and color attribution, demonstrating significant advantages of MaskSense in aligning with target preferences. See Appendix~\ref{C} for more qualitative comparisons.

\begin{table}[t]
\centering
\vspace{-10pt}
\begin{minipage}[t]{0.49\linewidth}
\centering
\footnotesize
\renewcommand{\tabcolsep}{1.5pt}
\caption{\textbf{Quantitative Comparison on the Human Preference
 Alignment Benchmark.} }
\label{table:hps}
\begin{tabular}{lcccc}
\toprule
\textbf{Model} & \textbf{HPS}$\uparrow$ & \textbf{DEQA}$\uparrow$ & \textbf{PicS.}$\uparrow$ & \textbf{ImgR.}$\uparrow$ \\
\midrule
    SD3.5-M~\cite{sd3.5} & 30.17 & 4.24 & 22.50 & 0.98  \\
    FLUX.1-dev~\cite{flux} & 31.13 & 4.37 & \textbf{22.97} & 1.06  \\
    Meissonic~\cite{bai2024meissonic} & 28.89 & 4.00 & 21.63 & 0.39  \\
    \quad$+$ GRPO~\cite{grpo} & 32.31 & 4.15 & 21.96 & 0.93 \\
    \quad$+$ MaskGRPO~\cite{maskgrpo} & 35.48 & 4.35 & 22.34 & 1.06 \\
    \quad$+$ MaskFocus~\cite{zhang2025maskfocus} & 35.52 & 4.39 & 22.39 & 1.09 \\
    \rowcolor{mygray}
    \textbf{\quad$+$ MaskSense (Ours)}  & \textbf{36.39} & \textbf{4.42} & 22.54 & \textbf{1.13} \\
    \rowcolor{mygray}
    \textcolor{deltagreen}{\quad$\Delta$ Over Base Model} & \inc{7.50} & \inc{0.42} & \inc{0.91} & \inc{0.74} \\
    \bottomrule
\end{tabular}
\end{minipage}\hfill
\begin{minipage}[t]{0.48\linewidth}
\centering
\footnotesize
\renewcommand{\tabcolsep}{1.5pt}
\renewcommand{\arraystretch}{1.48}
\caption{\textbf{Core Ablations.} PicS.: PickScore; ImgR.: ImageReward.}
\label{tab:core_ablation}
\begin{tabular}{lccccc}
\toprule
\textbf{Model} & \textbf{GenEval}$\uparrow$ &\textbf{HPS}$\uparrow$ & \textbf{DEQA}$\uparrow$ & \textbf{PicS.}$\uparrow$ & \textbf{ImgR.}$\uparrow$ \\
\midrule
Meissonic~\cite{bai2024meissonic} & 0.54 & 28.89 & 4.00 & 21.63 & 0.39 \\
+ GRPO~\cite{grpo} & 0.72 & 32.31 & 4.15 & 21.96 & 0.93 \\
w/o GETA & 0.79 & 32.62 & 4.20 & 22.09 & 0.95 \\
w/o SARS & 0.81 & 35.64 & 4.35 & 22.19 & 1.06 \\
\rowcolor{mygray}
\textbf{MaskSense} & \textbf{0.86} & \textbf{36.39} & \textbf{4.42} & \textbf{22.54} & \textbf{1.13} \\
\bottomrule
\end{tabular}
\end{minipage}
\vspace{-10pt}
\end{table}

\subsection{Analysis}
\textbf{Ablation Study.}
Tab.~\ref{tab:core_ablation} shows that removing either SARS or GETA consistently degrades performance across all metrics, confirming both components are indispensable to MaskSense.
As further evidenced by Fig.~\ref{fig:training}, the full model achieves more stable training dynamics and faster convergence throughout optimization, ultimately reaching a stronger performance.

\begin{table}[H]
\centering
\vspace{-4pt}
\footnotesize
\renewcommand{\tabcolsep}{2.4pt}
\caption{\textbf{Hyper-parameter Study}. We investigate the influence of the semantic strength $\lambda$, the temporal gate $\tau$ in the SARS, and the number of anchor steps $k$ in the GETA, respectively. }
\label{tab:param_ablation}
 \vspace{-6pt}
\begin{subtable}{0.34\textwidth}
    \centering
    \caption{Semantic Strength $\lambda$}
    \vspace{-3pt}
    \label{subtab:lambda}
    \begin{tabular}{lccccc}
    \toprule
    \textbf{Metric} & 0 & 0.3 & \textbf{0.5} & 0.7 & 1 \\
    \midrule
    HPS & 35.64 & 36.16 & 36.39 & 36.20 & 36.00 \\
    DEQA & 4.35 & 4.41 & 4.42 & 4.42 & 4.37 \\
    PickScore & 22.19 & 22.42 & 22.54 & 22.38 & 22.35 \\
    ImageReward & 1.06 & 1.12 & 1.13 & 1.04 & 1.06 \\
    \bottomrule
    \end{tabular}
\end{subtable}
\hspace{0.07\textwidth}
%
\begin{subtable}{0.33\textwidth}
    \centering
    \caption{Temporal Gate $\tau$}
    \vspace{-3pt}
    \label{subtab:tau}
    \begin{tabular}{ccccc}
    \toprule
    0 & 0.3 & \textbf{0.5} & 0.7 & 1 \\
    \midrule
    35.64 & 35.88 & 36.39 & 36.11 & 35.78 \\
    4.35 & 4.39 & 4.42 & 4.37 & 4.37 \\
    22.19 & 22.38 & 22.54 & 22.34 & 22.20 \\
    1.06 & 1.11 & 1.13 & 1.07 & 1.04 \\
    \bottomrule
    \end{tabular}
\end{subtable}
\hspace{-0.02\textwidth}
%
\begin{subtable}{0.22\textwidth}
    \centering
    \caption{Anchor Number $k$}
    \vspace{-3pt}
    \label{subtab:N}
    \begin{tabular}{@{\hspace{0.6em}}ccc@{\hspace{0.6em}}}
    \toprule
    4 & \textbf{6} & 8 \\ 
    \midrule
    35.79 & 36.39 & 36.18 \\
    4.36 & 4.42 & 4.39 \\
    22.37 & 22.54 & 22.50 \\
    1.07 & 1.13 & 1.11 \\
    \bottomrule
    \end{tabular}
\end{subtable}

\end{table}

\begin{figure}[t]
  \centering
  \begin{subfigure}{0.32\linewidth}
  \centering
    \includegraphics[width=\linewidth]{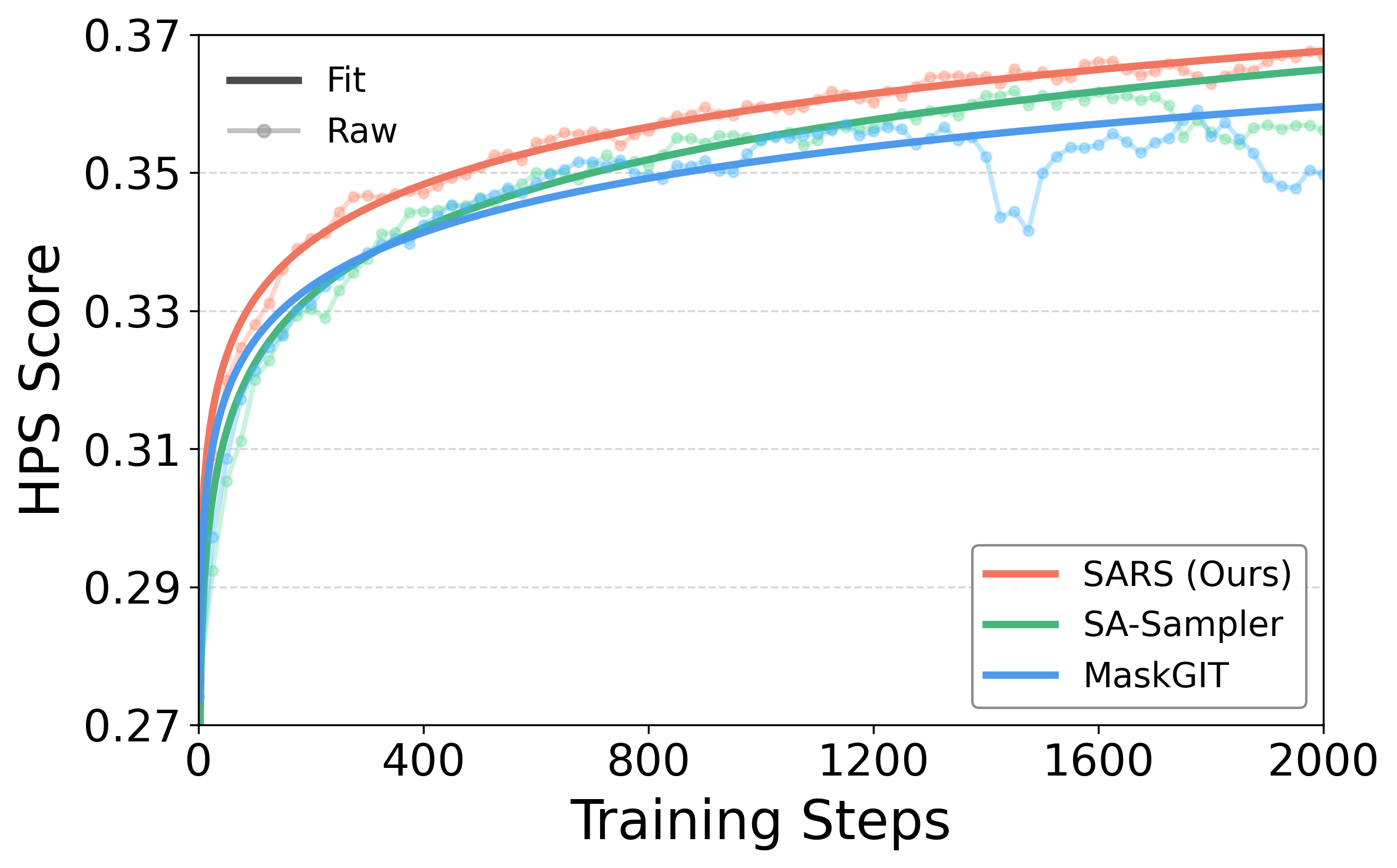}
    \caption{Effect of Sampling Strategy}
    \label{sampling}
  \end{subfigure}
  \hfill
  \begin{subfigure}{0.32\linewidth}
  \centering
    \includegraphics[width=\linewidth]{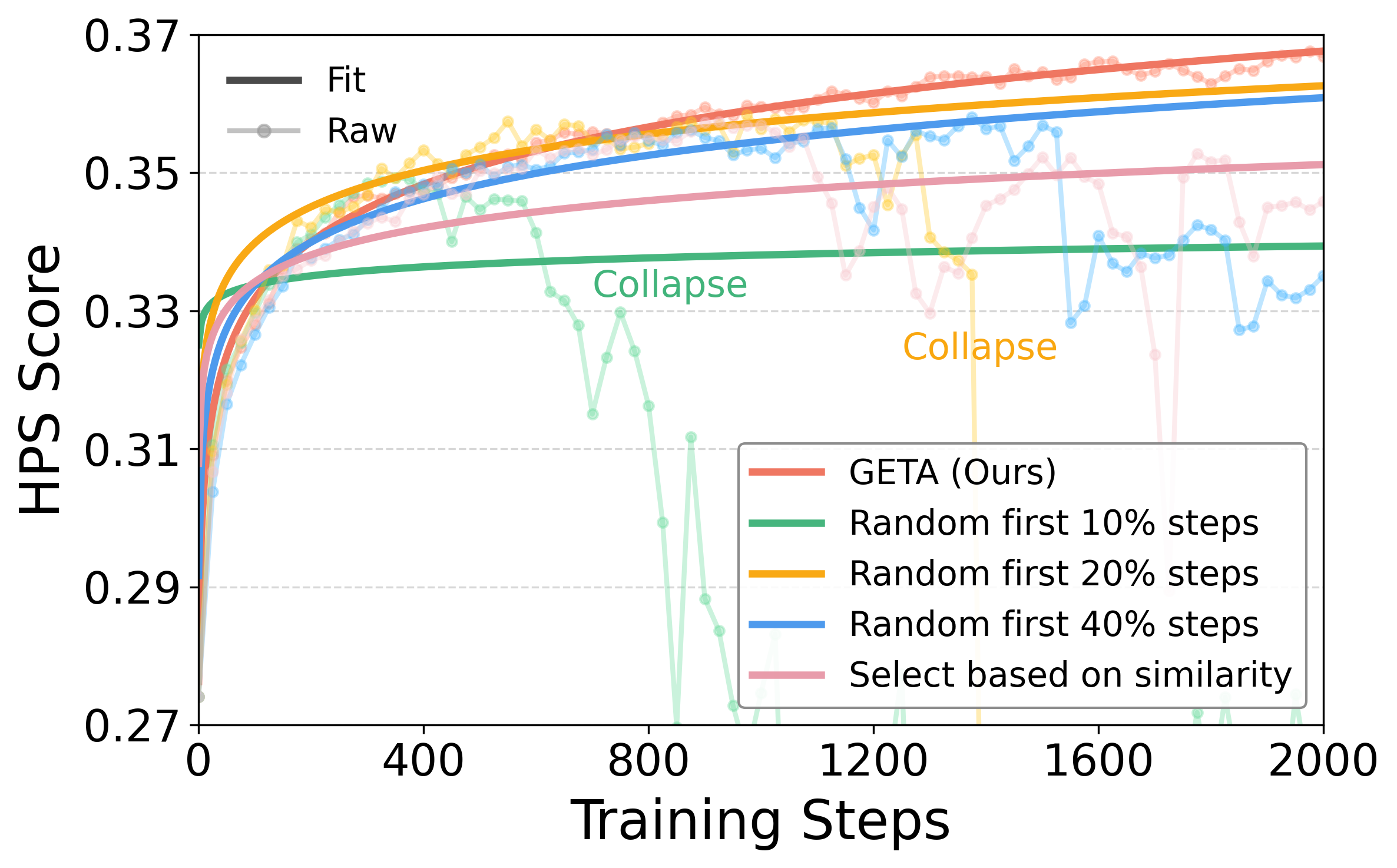}

    \caption{Effect of Anchoring Strategy}
    \label{select}
  \end{subfigure}
  \hfill
  \begin{subfigure}{0.32\linewidth}
  \centering
    \includegraphics[width=\linewidth]{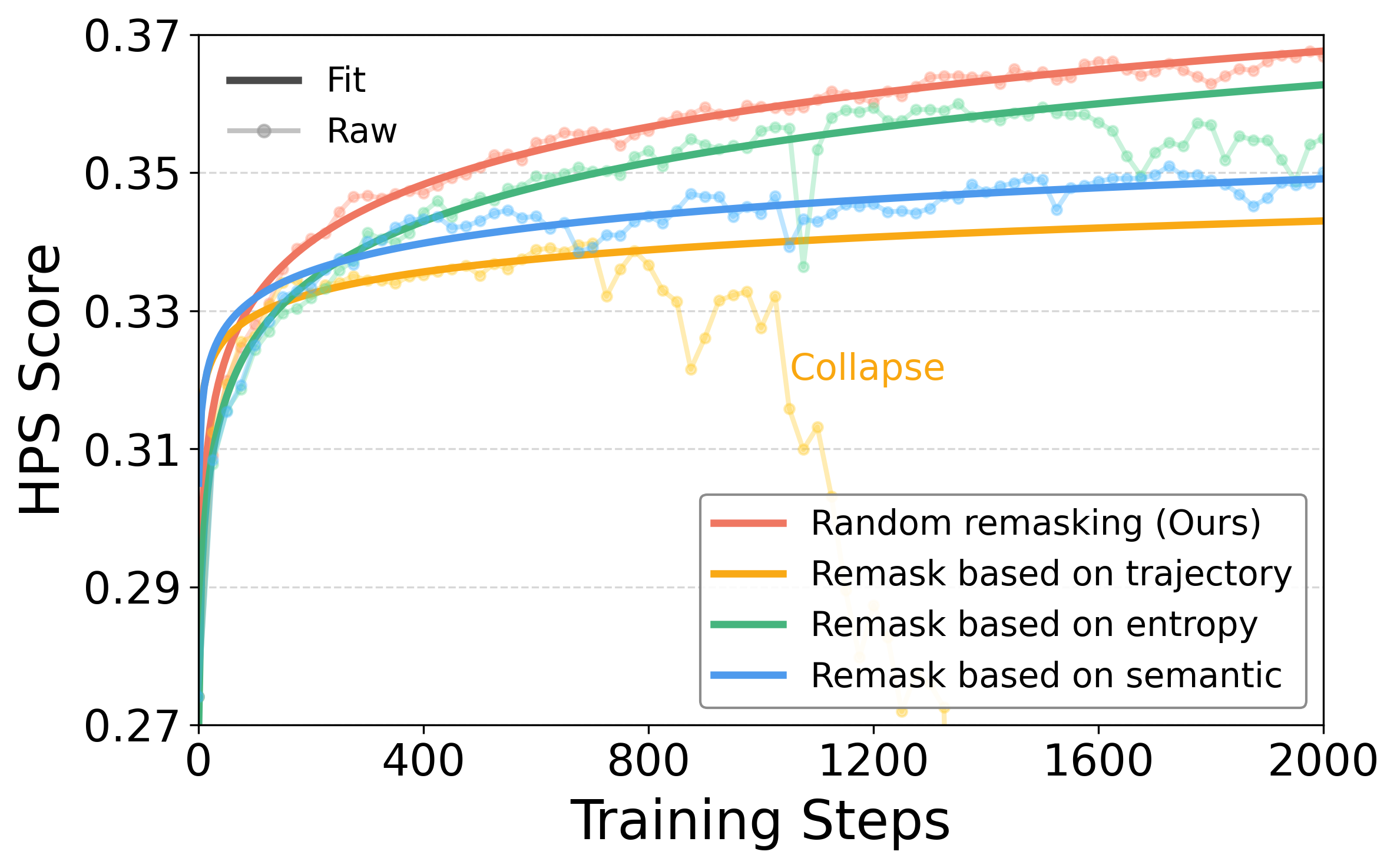}

    \caption{Effect of Remasking Strategy}
    \label{mask}
  \end{subfigure}
  \caption{\textbf{Ablation Analysis on Key Strategy Choices.} (a) SARS effectively balances exploration and exploitation, outperforming single-branch sampling strategies. (b) GETA anchors critical decision steps, enabling more principled policy optimization than heuristic alternatives. (c) Random remasking diversifies anchor states, producing more stable optimization.}
  \label{fig:ablation_analysis}
  \vspace{-2pt}
\end{figure}

\textbf{Effect of Semantic Anchored Routing Sampling.}
We compare SARS against MaskGIT and SA-Sampler alone in 
Fig.~\ref{fig:ablation_analysis} (a). SA-Sampler alone already improves 
over MaskGIT, confirming that anchoring subject tokens during 
early decoding stage effectively alleviates the visual exploration 
trap. However, the full SARS design brings a further and more 
stable gain, as semantic bias alone is insufficient without 
the dual-branch routing to mix exploratory and conservative 
trajectories, which amplifies reward contrast and sharpens 
advantage estimates. Moreover, Tab.~\ref{tab:param_ablation} (a) and (b) show that both semantic guidance strength $\lambda$ and temporal gate $\tau$ achieve optimal performance at 0.5.

\begin{figure}[t]
\centering
\vspace{-6pt}
\begin{overpic}[width=1\linewidth]{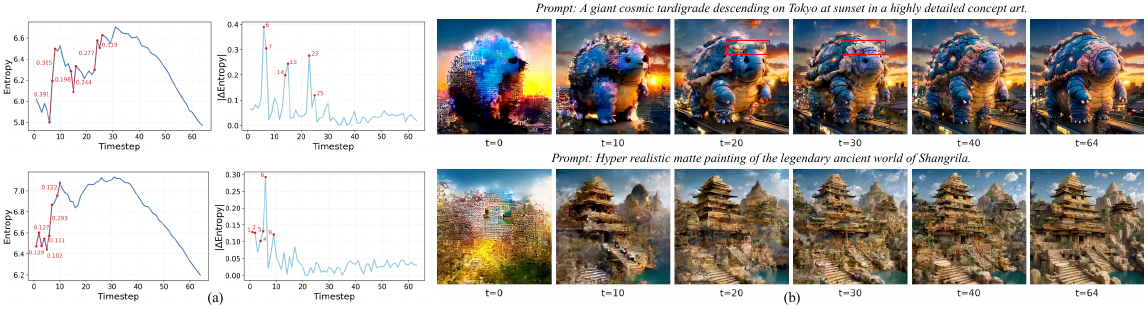}
\end{overpic}
\caption{\textbf{Anchored Timesteps of Diverse Prompts.} (a) Global entropy and absolute entropy difference between consecutive timesteps. (b) Visualization of generation process across timesteps.}
\label{entropy}
\end{figure}

\textbf{Effect of Cross-Branch Reward Contrast in Dual-Branch Routing.}
Jointly normalizing MaskGIT and SA-Sampler trajectories can introduce a branch-dependent component, motivating us to distinguish cross-branch contrast from trajectory-level quality.
Let $b_i\in\{\mathrm{MG},\mathrm{SA}\}$ denote the branch of rollout $i$, and let $\mu_b$ and $\mu$ be the branch-wise and pooled reward means. The centered reward decomposes as $R_i-\mu=(R_i-\mu_{b_i})+(\mu_{b_i}-\mu)$, where the two terms capture within-branch trajectory quality and cross-branch reward contrast, respectively.
We remove the latter mean-offset term using branch-conditioned normalization within each $G=8$ rollout group, which enforces
$\mathbb{E}\!\left[\hat{A}_i \mid b_i=\mathrm{SA}\right]=\mathbb{E}\!\left[\hat{A}_i \mid b_i=\mathrm{MG}\right]
=0.$ 
Dual-branch training still outperforms MaskGIT-only $+$ GETA (36.08 vs.\ 35.64 HPS), while pooled normalization reaches 36.39, showing that cross-branch contrast contributes to, but does not account for, the overall gain.
A 300-prompt overlap analysis further supports this conclusion: the branch reward ranges overlap for 90.7\% of prompts, with 26.9\% of SA-Sampler rollouts below the pooled mean and 25.7\% of MaskGIT rollouts above it. Thus, sampler identity does not determine the advantage sign. 
Since all variants use MaskGIT for evaluation, the observed gain stems from the learned policy itself.

\textbf{Effect of Global Entropy Transition Anchoring.}
We compare GETA against random early-step selection and 
similarity-based selection~\cite{zhang2025maskfocus} in 
Fig.~\ref{fig:ablation_analysis} (b). Although restricting anchors to the first 10\% or 20\% of steps brings faster initial gains, it ultimately leads to training collapse, possibly due to highly masked states inducing unstable optimization targets.
Similarity-based selection appears more stable initially but exhibits increasing instability in later stages, as embedding similarity captures visual representational change but fails to align with the model's most decisive policy decision points.
As shown in Fig.~\ref{entropy}, the top row exhibits dominant transitions spread across the first 30 steps where each committed decision critically shapes the overall image structure, while the bottom row concentrates transitions within the first 10 steps with later steps contributing only minor detail refinement. This prompt-adaptive distribution of critical steps validates the rationality of GETA's entropy-guided anchor selection.
Besides, Tab.~\ref{tab:param_ablation} (c) demonstrates that $k=6$ achieves the best trade-off between optimization focus and trajectory coverage under the same GPU hours.


\textbf{Effect of Random Remasking.}
We conduct a systematic study of remasking strategies 
for anchor-state reconstruction in Fig.~\ref{fig:ablation_analysis} (c), as masked positions directly affect anchor-state probability estimation.  Random remasking achieves the best convergence and stability, while trajectory-based remasking leads to early collapse. Entropy-guided and semantics-guided remasking remain trainable yet consistently underperform, indicating that randomizing masked positions while preserving the mask budget produces greater anchor-state diversity and more robust optimization.

\textbf{Discussion of Reward Hacking.}
Beyond task-specific metrics, we assess reward hacking 
via the DrawBench~\cite{drawbench} metrics reported in Tab.~\ref{table:hps}. 
MaskSense demonstrates outstanding alignment capability, indicating that the model generalizes well beyond the 
training reward without overfitting to optimized metrics. 
Furthermore, we present the joint multi-reward training 
curves of GenEval and CLIP~\cite{clip} rewards in Fig.~\ref{fig:training} (c), which remain consistently stable with superior performance throughout training. These advancements demonstrate the strong robustness of the proposed MaskSense.

\begin{table}[t]
\centering
\vspace{-4pt}
\scriptsize
\captionsetup{skip=3pt}
\caption{\textbf{Transfer to Show-o on HPS.} Evaluation HPS scores and cumulative training GPU hours.}
\label{tab:showo_transfer}
\resizebox{\linewidth}{!}{
\begin{tabular}{lccccccccccc}
\toprule
\textbf{Step} & 0 & 200 & 400 & 600 & 800 & 1K & 1.2K & 1.4K & 1.6K & 1.8K & 2K \\
\midrule
Mask-GRPO HPS & 26.54 & 27.28 & 28.09 & 28.54 & 29.05 & 29.44 & 29.56 & 29.72 & 29.74 & 29.72 & 29.81 \\
\rowcolor{mygray}
MaskSense HPS & 26.54 & 27.60 & 28.22 & 28.92 & 29.30 & 29.66 & 29.82 & 29.98 & 29.97 & 30.05 & 30.09 \\
Mask-GRPO GPU Hours & 0.00 & 11.10 & 22.17 & 33.21 & 44.30 & 55.37 & 66.44 & 77.51 & 88.61 & 99.66 & 110.72 \\
\rowcolor{mygray}
MaskSense GPU Hours & 0.00 & 11.45 & 22.89 & 34.31 & 45.73 & 57.15 & 68.56 & 79.96 & 91.37 & 102.77 & 114.18 \\
\bottomrule
\end{tabular}}

\vspace{2pt}
\footnotesize
\renewcommand{\tabcolsep}{4.2pt}
\caption{\textbf{VQA Evaluation on GenEval.} Results cover all prompts with four images per prompt.}
\label{tab:vqa_geneval}
\resizebox{\linewidth}{!}{
\begin{tabular}{lccccccc}
\toprule
\textbf{Model} & \textbf{Overall}$\uparrow$ & \textbf{Single Obj.}$\uparrow$ & \textbf{Two Obj.}$\uparrow$ & \textbf{Counting}$\uparrow$ & \textbf{Color}$\uparrow$ & \textbf{Position}$\uparrow$ & \textbf{Color Attr.}$\uparrow$ \\
\midrule
Meissonic & 0.7006 & 0.9750 & 0.8030 & 0.5687 & 0.9149 & 0.6400 & 0.5925 \\
$+$ GRPO & 0.8616 & \textbf{0.9875} & 0.9343 & 0.9109 & 0.9335 & 0.8233 & 0.7700 \\
\rowcolor{mygray}
\textbf{$+$ MaskSense} & \textbf{0.9189} & 0.9844 & \textbf{0.9432} & \textbf{0.9234} & \textbf{0.9362} & \textbf{0.9567} & \textbf{0.8662} \\
\rowcolor{mygray}
\textcolor{deltagreen}{$\Delta$ Over Base Model}
& \inc{0.2183} & \inc{0.0094} & \inc{0.1402}
& \inc{0.3547} & \inc{0.0213} & \inc{0.3167}
& \inc{0.2737} \\
\bottomrule
\end{tabular}}

\vspace{2pt}
\footnotesize
\renewcommand{\tabcolsep}{5pt}
\caption{\textbf{Human Preference Evaluation on HPS.} Image order is randomized in each comparison.}
\label{tab:human_preference}
\begin{tabular}{lcccc}
\toprule
\textbf{Comparison} & \textbf{Dimension} & \textbf{Opponent Win} & \textbf{MaskSense Win} & \textbf{Tie} \\
\midrule
Meissonic vs.\ MaskSense & Aesthetic & 28 (7\%) & \textbf{348 (87\%)} & 24 (6\%) \\
Meissonic vs.\ MaskSense & Semantic & 24 (6\%) & \textbf{92 (23\%)} & 284 (71\%) \\
GRPO vs.\ MaskSense & Aesthetic & 84 (21\%) & \textbf{216 (54\%)} & 100 (25\%) \\
GRPO vs.\ MaskSense & Semantic & 32 (8\%) & \textbf{60 (15\%)} & 308 (77\%) \\
\bottomrule
\end{tabular}
\vspace{-5pt}
\end{table}

\textbf{Model Transferability and Independent Evaluations.}
\textit{Transfer to Show-o.}
We integrate SARS and GETA into the official Mask-GRPO implementation on Show-o~\cite{show-o} while retaining its training configuration. As shown in Tab.~\ref{tab:showo_transfer}, MaskSense consistently outperforms Mask-GRPO and reaches 30.09 HPS at 2K steps, versus 29.81 for Mask-GRPO and 26.54 for the pretrained model, with only 3.1\% additional GPU hours. This demonstrates transfer beyond Meissonic~\cite{bai2024meissonic} with modest overhead.

\textit{Independent Evaluations.}
Using Qwen-VL-Plus~\cite{qwen3vl} as an independent VQA evaluator, we assess object existence, counting, color, and spatial relations on the GenEval test set. 
Tab.~\ref{tab:vqa_geneval} shows that MaskSense achieves 91.89\% overall accuracy, outperforming GRPO (86.16\%) and Meissonic (70.06\%). 
We further conduct human pairwise comparisons on HPS prompts. As shown in Tab.~\ref{tab:human_preference}, MaskSense is aesthetically preferred over Meissonic and GRPO in 87\% and 54\% of cases, respectively, while also leading in semantic alignment despite frequent ties. These reward-independent evaluations confirm improvements in both compositional faithfulness and perceptual quality.
  
\section{Conclusion}
We present MaskSense to address the visual exploration trap in masked generative models, where confidence-based sampling prematurely suppresses subject-token diversity and undermines the effectiveness of GRPO optimization. 
By anchoring subject tokens during early decoding via 
semantic priors, MaskSense preserves the high-entropy exploration space that confidence-based sampling would otherwise collapse, enabling sharper intra-group advantage estimation through dual-branch rollout routing.
Building on the enriched rollout diversity, MaskSense further identifies timesteps of maximal entropy transition as optimization anchors to address the uneven 
contribution of individual steps to policy learning, concentrating gradient updates where token commitment is most decisive.
%
Extensive experiments demonstrate the gains achieved by MaskSense and validate the critical role of confronting the visual exploration trap for RL optimization in masked generative models.  
\section*{Acknowledgments}
This work was supported in part by the National Natural Science Foundation of China (NSFC) under Grants Grants 62225207, 62436008, 62422609, 62306295, and 62576328, in part by the CAS Project for Young Scientists in Basic Research under Grant YSBR-122, and in part by the Fundamental Research Funds for the Central Universities under Grant WK2102026001. 

\bibliographystyle{plain}
\bibliography{ref}

\appendix
\newpage
\section{Details on the Experimental Setup}
\label{A}
\subsection{Dataset and Reward}
\label{A.1}
\textbf{Dataset.} For the compositional image generation task, 50K prompts are constructed using the official GenEval~\citep{ghosh2023geneval} scripts, following Flow-GRPO~\cite{flow-grpo}, covering six sub-tasks with the prompt sampling ratio set to Position : Counting : Attribute Binding : Colors : Two Objects : Single Object = 7 : 5 : 3 : 1 : 1 : 0. The same dataset is also used for multi-reward joint optimization. For human preference alignment, we use 10K training samples from the HPS v2 dataset~\citep{hps}, a large-scale benchmark collecting pairwise human judgments over generated images across diverse prompt styles. We further evaluate out-of-distribution generalization on DrawBench~\citep{drawbench}, a benchmark comprising 200 prompts across 11 diverse categories that probes broad generative capabilities beyond the training distribution.

\textbf{Reward.}
For compositional image generation, we adopt the GenEval reward, which is computed based on rule-based detection of object presence, count, spatial position, and color attribution. For human preference alignment, we employ the HPS~\citep{hps} reward model. For multi-reward joint optimization, we simultaneously optimize GenEval and CLIP~\citep{clip} rewards.

\subsection{Implementation Details}
\label{A.2}
\textbf{Training.}
All experiments are conducted on Meissonic~\citep{bai2024meissonic}, a masked text-to-image model at $1024 \times 1024$ resolution, trained on 8 NVIDIA A100 GPUs. We use the AdamW optimizer with a learning rate of $2\times 10^{-6}$. The maximum prompt length and completion length are set to 512 and 1024, respectively, with a sampling temperature of $1.0$. For GRPO optimization, we set the group size to $G=8$, with a per-device batch size of 1 and gradient accumulation steps of 1. Following the standard masked-generation RL setup~\citep{zhang2025maskfocus}, we adopt a classifier-free guidance scale of 5 during both training and inference, and perform 64 iterative decoding steps. The number of selected optimization anchors is set to $k=6$, and the KL penalty coefficient is set to $\beta=0.01$. For SARS, the semantic guidance strength is set to $\lambda=0.5$ and the temporal gate threshold to $\tau=0.5$.

\textbf{Evaluation.} For compositional image generation, we evaluate on the GenEval benchmark~\citep{ghosh2023geneval} using its official evaluation pipeline, which detects object bounding boxes and colors and infers their spatial relations. The evaluation set comprises 553 prompts, for each of which we generate 4 images to ensure stable and reliable score estimation. For human preference alignment, we evaluate on a held-out test set of 400 prompts from HPS v2~\citep{hps}. For out-of-distribution generalization, we evaluate on DrawBench~\citep{drawbench}, a benchmark of 200 prompts spanning 11 diverse categories, generating 5 images per prompt to ensure broader coverage and stable generalization assessment. 
To ensure fairness and stability during evaluation, all models are assessed using the original MaskGIT sampler as the decoding strategy at evaluation time, regardless of the sampling strategy employed during training. The proposed SA-Sampler is exclusively utilized during the RL training phase to promote exploratory diversity in rollout generation and is not applied at evaluation time.
All quantitative results reported in the main paper are obtained under the above evaluation protocol. To ensure statistical reliability, all metrics are averaged over multiple generated images per prompt, which reduces variance and yields more stable estimates across different random seeds, ensuring that the reported numbers reflect the model's generative capability rather than individual sampling artifacts.

\subsection{Quality Metrics}
\label{A.3}
The details of task accuracy metrics and quality metrics are as follows:
\begin{itemize}
\item GenEval~\citep{ghosh2023geneval}: An object-focused evaluation framework that benchmarks T2I models on six compositional image generation tasks. The official pipeline detects object bounding boxes and colors and infers spatial relations between them. Rewards follow rule-based criteria: (i) Counting: $r=1-|N_{gen}-N_{ref}|/N_{ref}$; (ii) Position / Color: a partial reward is granted when the object count is correct, with the remainder awarded upon correct prediction of position or color.

\item HPS~\citep{hps}: A scoring model obtained by fine-tuning CLIP~\citep{clip} on the Human Preference Dataset v2 (HPD v2), a large-scale collection of 798K pairwise human preference annotations over images generated from diverse prompts and models. HPS v2 predicts the likelihood that a generated image will be preferred by humans, serving as a reliable proxy for aesthetic quality and text-image alignment.

\item DeQA~\citep{deqa}: A multimodal large language model-based image quality assessment metric that estimates overall perceptual quality by accounting for distortions, texture degradation, and AI-specific artifacts, trained via a fidelity loss over soft score distributions.

\item PickScore~\citep{pickscore}: A preference scoring model fine-tuned from CLIP~\citep{clip} on Pick-a-Pic, a large-scale dataset of real user preference choices between pairs of images generated from identical prompts, providing judgments strongly aligned with human evaluation across text-image alignment and visual aesthetics.

\item ImageReward~\citep{imagereward}: The first general-purpose reward model for text-to-image generation, trained to capture human preference dimensions encompassing text-image alignment, perceptual fidelity, and harmlessness.
\end{itemize}

\begin{algorithm*}[t]
\caption{MaskSense Training Process}
\label{alg:masksense}
\begin{algorithmic}[1]
\Require Reward model $\mathrm{RM}$; Prompt dataset $\mathcal{C}$; Decoding steps $T$; Anchored steps $\mathcal{K}$; Group size $G$.
\Ensure Optimized policy model parameters $\theta$
\State Initialize policy model $\pi_{\theta}$
\While{not converged}
    \State Sample batch of prompts $\mathcal{C}_b \sim \mathcal{C}$
    \State Update old policy model: $\pi_{\theta_{\text{old}}} \leftarrow \pi_{\theta}$
     \State \Comment{\textit{// Sampling Stage}}
    \For{each prompt $c \in \mathcal{C}_b$}
        \For{generating $i$-th image from $i = 1$ \textbf{to} $G$}
            \For{decoding timestep $t = 0$ \textbf{to} $T-1$}
                \State Compute semantic prior $\mathbf{s}_t$ in Eq.~\ref{eq7} and retention score $\hat{\mathbf{p}}_t$ in Eq.~\ref{eq9}
                \State Retain tokens to obtain $x_{t+1}^{i}$ via assigned branch sampler
            \EndFor
            \State Compute group-averaged entropy $H_t$ in Eq.~\ref{eq10} and advantage $\hat{A}^{i}_t$ in Eq.~\ref{eq3}
        \EndFor
        \State \Comment{\textit{// Optimization Stage}}
        \State Compute entropy transition magnitudes $\Delta H_t$ and select anchor set $\mathcal{K}$ in Eq.~\ref{eq11}
        \For{each anchor step $t \in \mathcal{K}$}
            \State Sample random remasking $M^{\text{rand}}_t$ with same budget as $M^{\text{traj}}_t$ in Eq.~\ref{eq12}
            \State Construct remasked input $\tilde{x}_t$ from final latent $x_T$ in Eq.~\ref{eq13}
            \State Compute policy optimization objective $\mathcal{J}_{\text{MaskSense}}(\theta)$ in Eq.~\ref{eq14}
        \EndFor
    \EndFor
\EndWhile
\end{algorithmic}
\end{algorithm*}

\section{Limitations and Societal Impacts}
\label{B}
\textbf{Limitations.} While our method demonstrates strong performance across multiple benchmarks, several limitations remain. First, the proposed reinforcement learning framework is grounded in the general MaskGIT paradigm, which is adopted by the vast majority of existing masked generative models as their core sampling strategy. It is therefore in principle compatible with a broad family of such models. Validating and extending its application to a wider range of base architectures~\cite{yang2025mmada} will be one of the primary focuses of our future work. 
Second, the current work focuses solely on image generation, leaving the exploration of reinforcement learning for multi-modal discrete diffusion models to future work, including the analysis of sampling and optimization across different modalities toward a unified RL framework for multi-modal masked generative models~\cite{luminadimoo}.

\textbf{Societal Impacts.} This work aims to advance text-to-image generation by aligning masked generative models with human preferences and compositional understanding. The proposed framework has the potential to benefit a wide range of creative, educational, and industrial applications. However, improved generative capabilities may also be exploited to produce misleading or harmful visual content, raising concerns about misinformation and malicious use. Additionally, as the model is trained on large-scale data, it may inadvertently inherit and amplify societal biases present in the training corpus. We encourage responsible deployment of the proposed method and hope to inspire future research on safety, fairness, and interpretability in generative modeling.

\section{Additional Experiments}
\label{C}
\begin{figure}[t]
\centering
\begin{overpic}[width=1\linewidth]{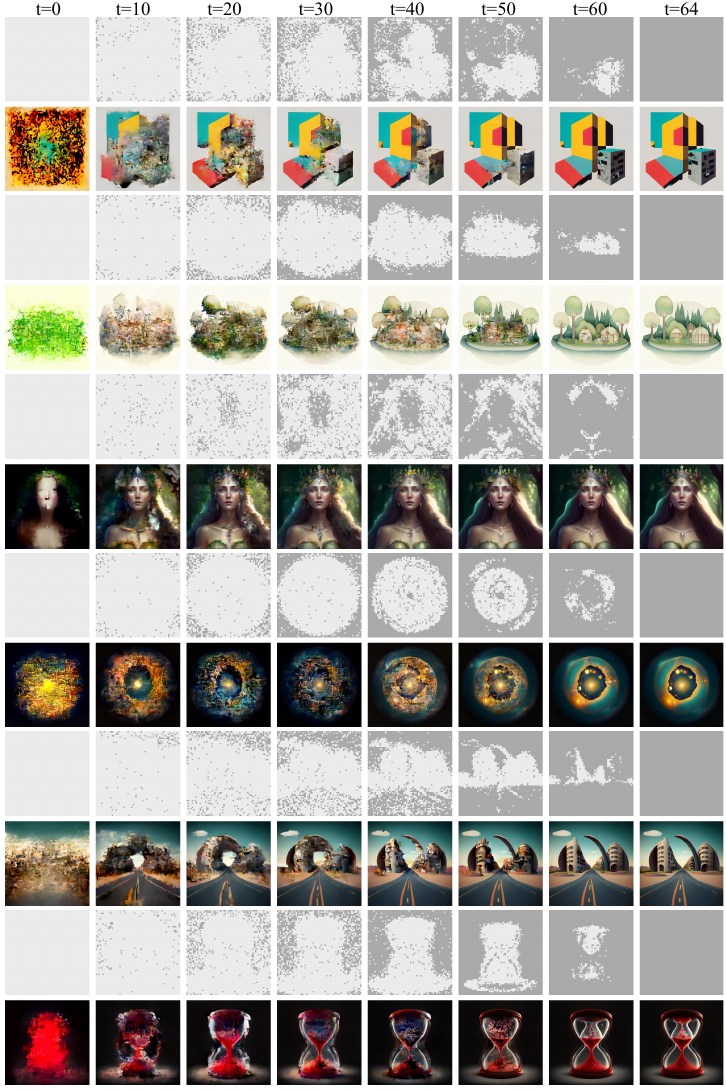}
\end{overpic}
\caption{
\textbf{Sampling Process Visualization of MaskGIT Sampler on Meissonic baseline.}
}
\label{fig:obs_maskgit}
\end{figure}

\begin{figure}[t]
\centering
\begin{overpic}[width=1\linewidth]{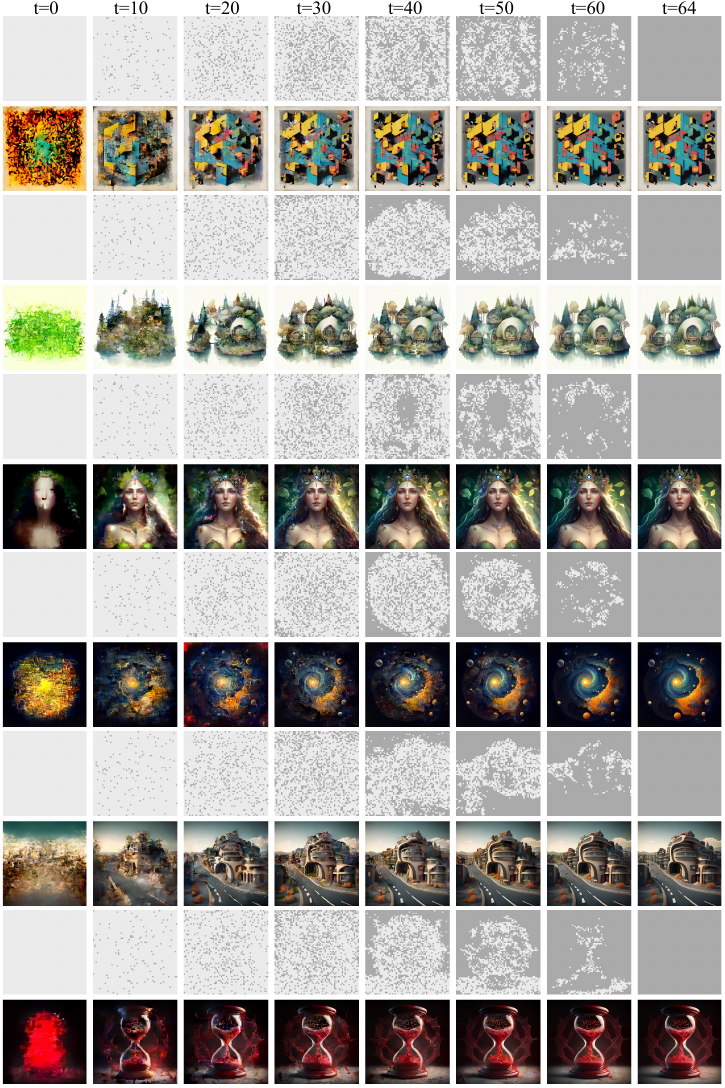}
\end{overpic}
\caption{
\textbf{Sampling Process Visualization of SA-Sampler on Meissonic baseline.}
}
\label{fig:obs_sasampler}
\end{figure}

\begin{figure}[t]
\centering
\begin{overpic}[width=1\linewidth]{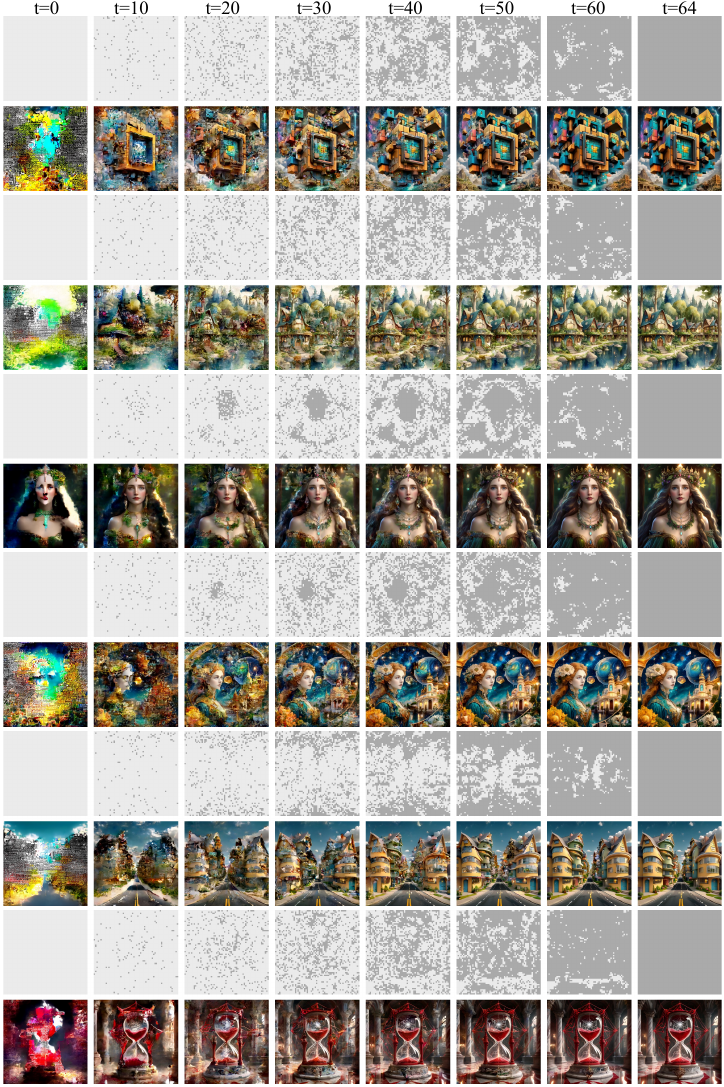}
\end{overpic}
\caption{
\textbf{Sampling Process Visualization of MaskGIT Sampler on MaskSense after RL Training.}
}
\label{fig:obs_masksense}
\end{figure}

\textbf{Observation Discussion.}
Fig.~\ref{fig:obs_maskgit} and~\ref{fig:obs_sasampler} illustrate the sampling trajectories of the MaskGIT sampler and SA-Sampler across diverse prompt categories, including abstract art, natural landscapes, character portraits, architectural scenes, and still-life compositions.
A consistent pattern emerges across all categories: the MaskGIT sampler systematically resolves background regions in the early decoding steps while deferring subject tokens to later stages, confirming that the visual exploration trap is a pervasive structural bias rather than a prompt-specific artifact.
In contrast, the SA-Sampler, without any modification to the model parameters, redirects early token commitment toward semantically critical subject regions, enabling the discovery of higher-quality generative trajectories through broader subject-space exploration.
Fig.~\ref{fig:obs_masksense} further reveals how RL training reshapes the decoding behavior of the model. While the pre-training MaskGIT sampler follows a rigid background-to-subject commitment order, the MaskSense trained model exhibits a markedly more flexible and diverse sampling process, with subject tokens being resolved earlier and more consistently across different prompts.
This behavioral shift demonstrates that MaskSense not only improves final image quality but fundamentally restructures the model's generative strategy toward a more diverse and flexible sampling process.

The corresponding prompts in Fig.~\ref{fig:obs_maskgit},~\ref{fig:obs_sasampler},~\ref{fig:obs_masksense} are: 
(1) A spray painted and analogue collage with canvas texture in a contemporary art style featuring a mathematically correct Tetris design. 
(2) A town of pod homes integrated in a forest area with water and trees, depicted in a detailed watercolor by Lurid. 
(3) An oil painting portrait of a beautiful dryad wearing an ombre velvet gown, with long hair and a tiara, adorned with dozens of jeweled necklaces, and illuminated with dramatic cinematic lighting. 
(4) A digital painting of a small universe with intricate details and ornate features, featuring the styles of Claude Monet and Vincent van Gogh. 
(5) A lot of building on each side of the road, with a very curvy road in the middle. 
(6) A red cobweb is seen inside a marble with an hourglass, lightning and intricate details, creating a sense of awe with swirling mist.

\begin{figure}[t]
  \centering

  \begin{subfigure}{0.32\linewidth}
  \centering
    \includegraphics[width=\linewidth]{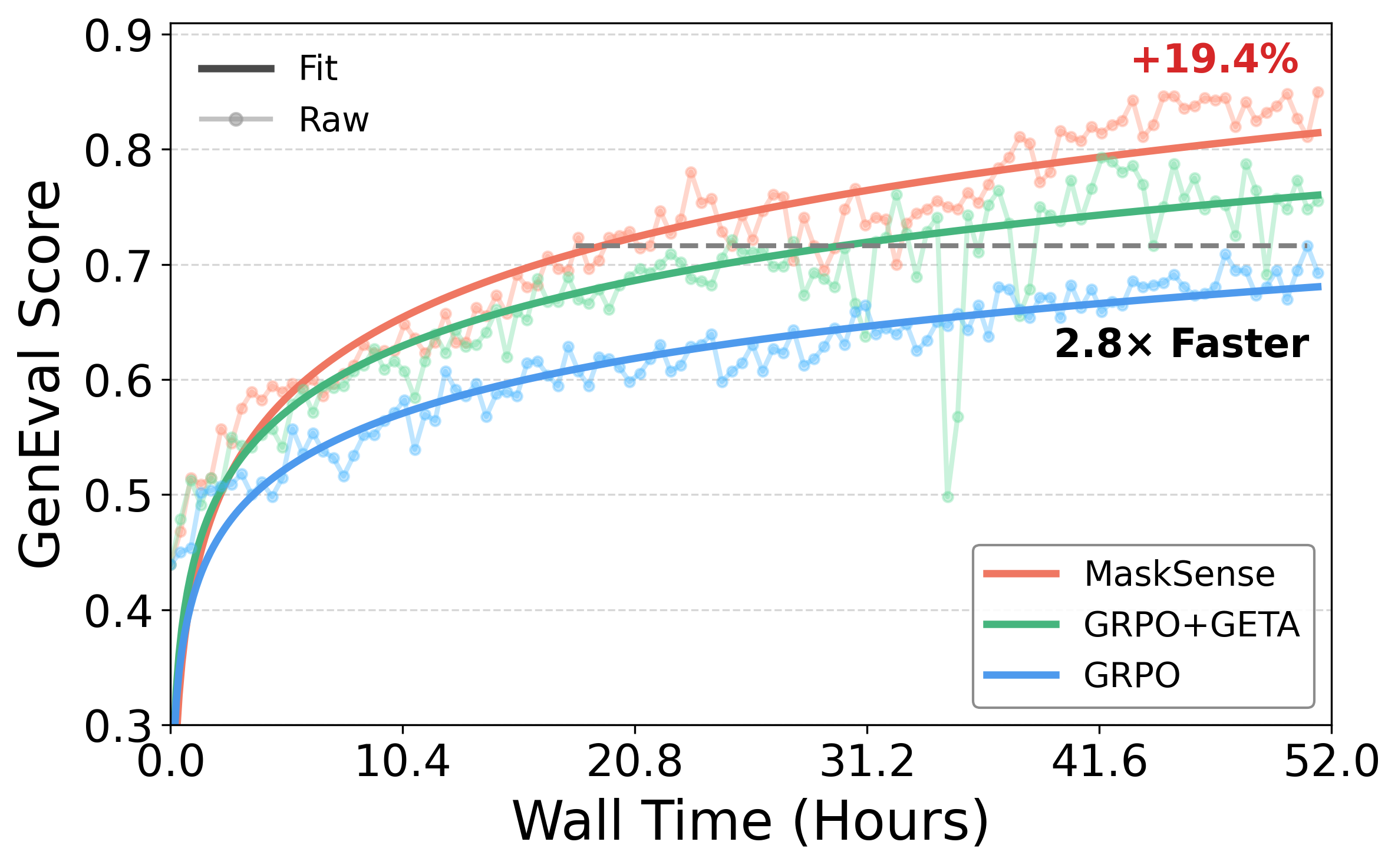}
    \caption{Compositional Image Gen.}
    \label{fig:geneval_app}
  \end{subfigure}
  \hfill
  \begin{subfigure}{0.32\linewidth}
  \centering
    \includegraphics[width=\linewidth]{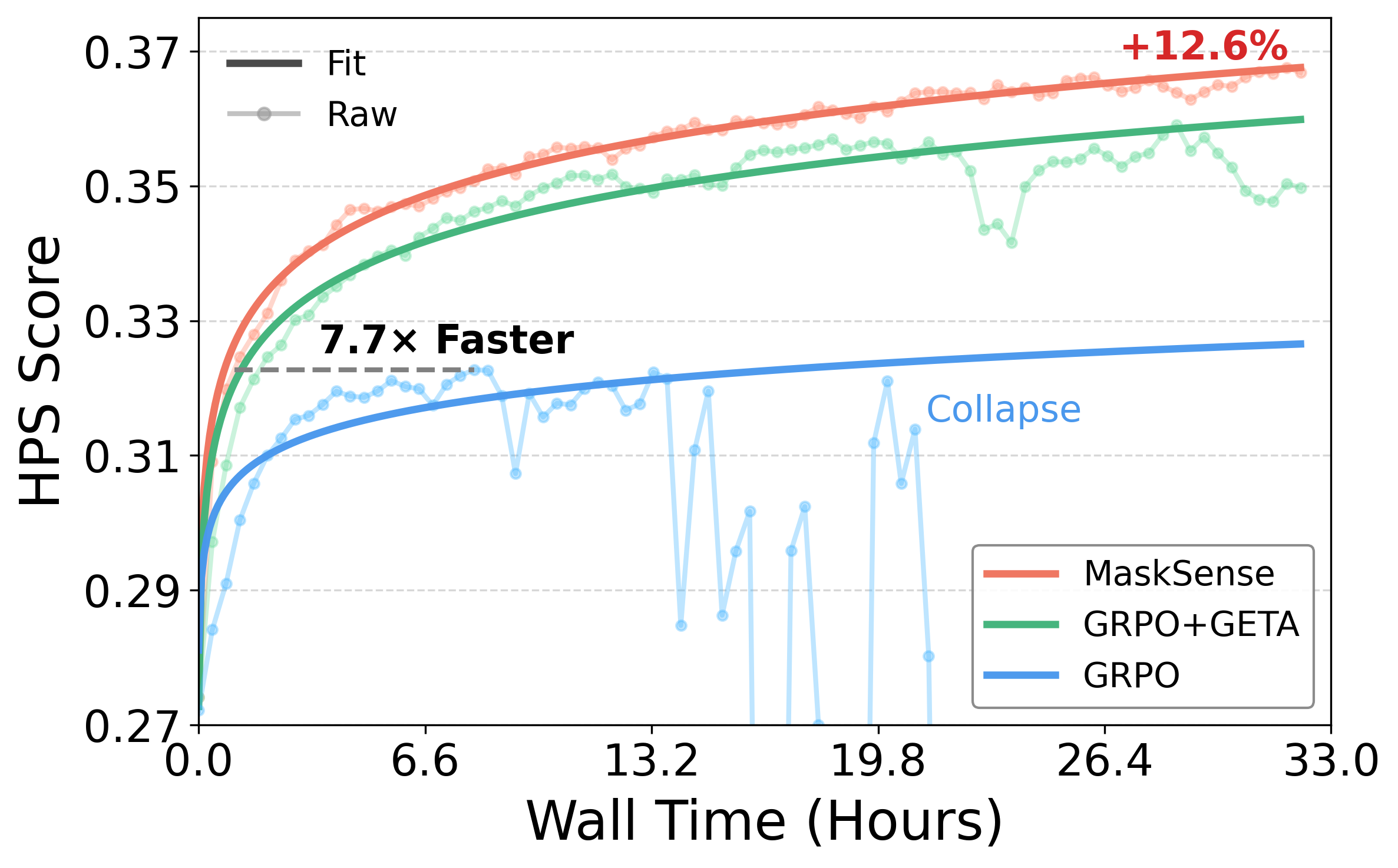}
    \caption{Human Preference Alignment}
    \label{fig:hps_app}
  \end{subfigure}
  \hfill
  \begin{subfigure}{0.32\linewidth}
  \centering
    \includegraphics[width=\linewidth]{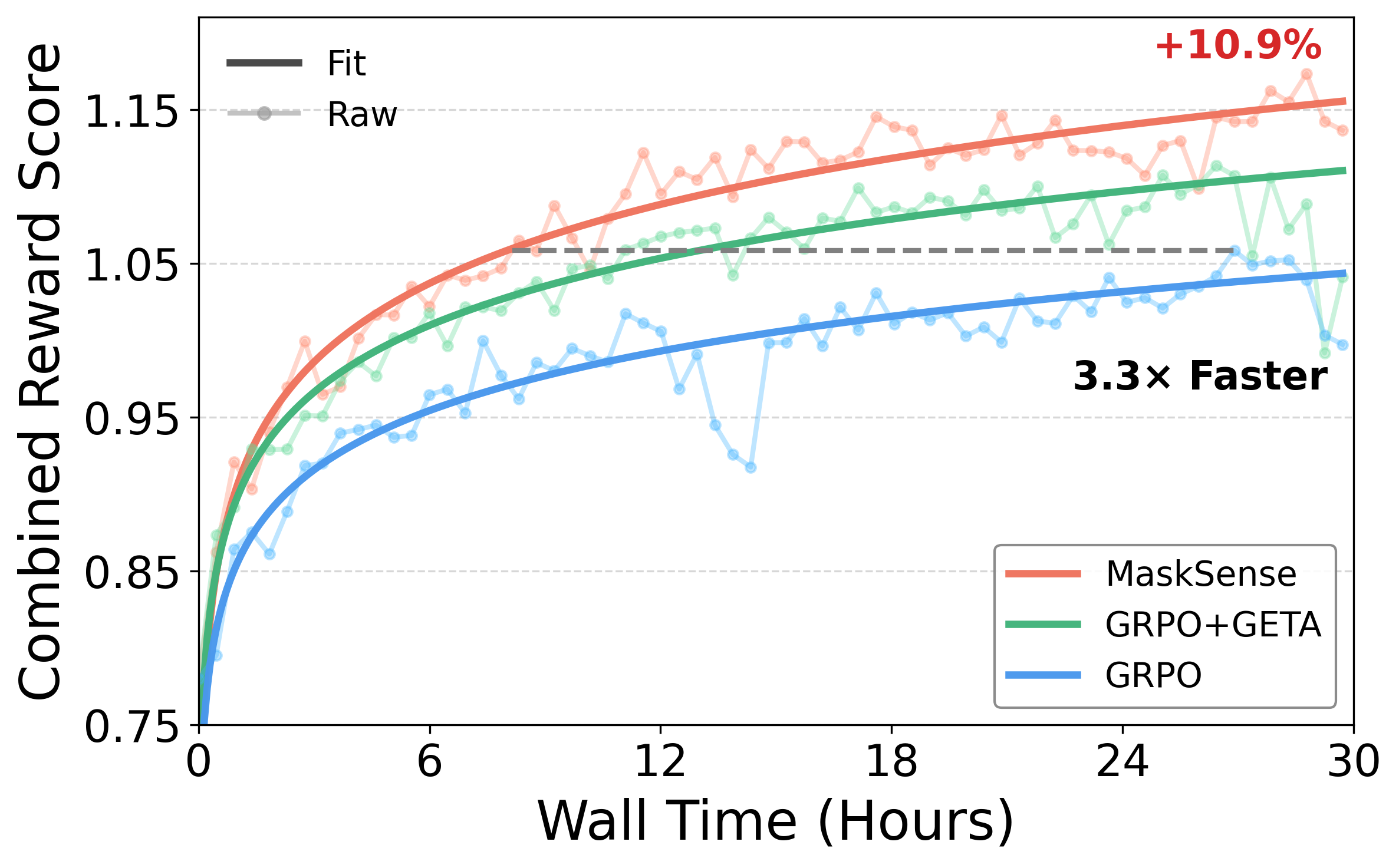}
    \caption{Multi-Reward Optimization}
    \label{fig:multi-reward_app}
  \end{subfigure}
  \caption{\textbf{Comparison of Compute Overhead Dynamics.} Fig.(a) to (c) refer to the tasks of compositional image generation, human preference alignment, and multi-reward optimization, respectively.}
  \label{fig:gpu_hours}

\end{figure}

\textbf{Computational Efficiency.}
Although MaskSense introduces additional operations for semantic prior extraction in the SA-Sampler and global entropy computation in GETA, the resulting computational overhead is negligible in practice.
As shown in Fig.~\ref{fig:gpu_hours}, the GPU hours curve of MaskSense remains fully consistent with the training dynamics curves in Fig.~\ref{fig:training} of the main paper at the same convergence speed. Taking the human preference alignment task as an example, MaskSense incurs an additional overhead of only 0.064 GPU hours per 100 training steps compared to the baseline. This marginal cost is negligible relative to the substantial performance gains achieved by MaskSense, demonstrating that our method improves generation quality with virtually no additional computational burden. The trend curves are obtained by robust logarithmic fitting $y = a + b\log(1+x)$ with outlier rejection.

\begin{figure}[t]
\centering
\begin{overpic}[width=0.85\linewidth]{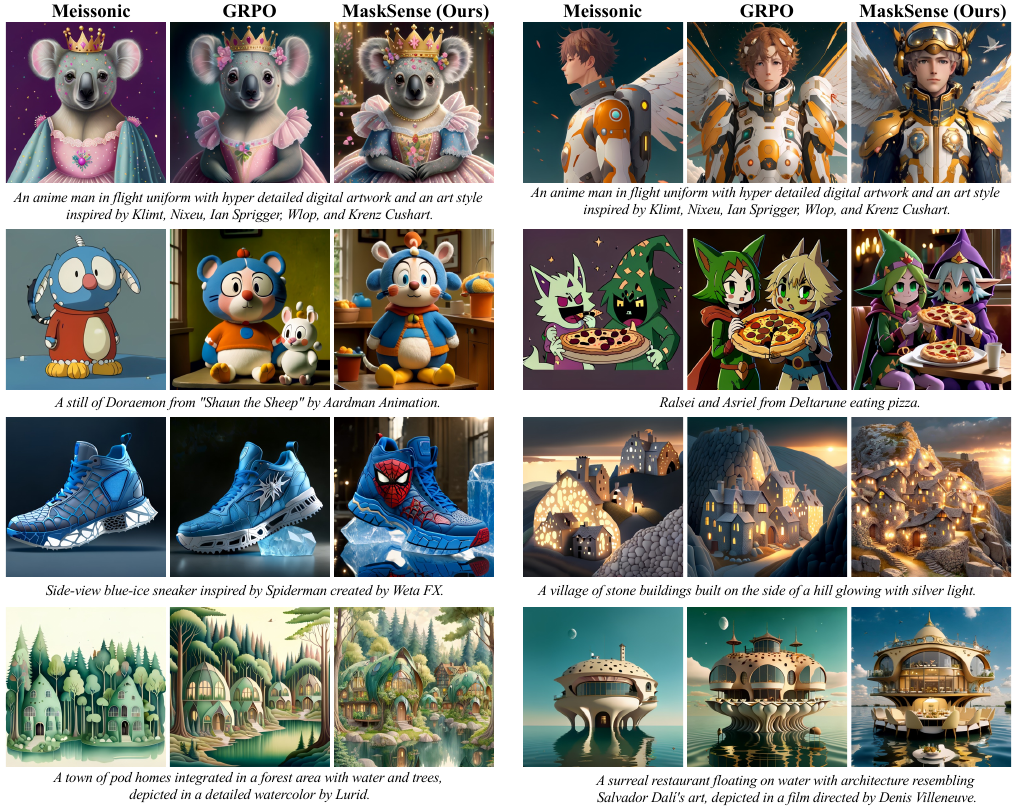}
\end{overpic}
\caption{\textbf{Qualitative Comparisons on Human Preference Alignment.} MaskSense generates images with richer stylistic details, superior texture fidelity, and more faithful rendering of complex artistic styles compared to Meissonic and GRPO baselines.}
\label{fig:supp_hps}
\end{figure}

\begin{figure}[t]
\centering
\begin{overpic}[width=0.85\linewidth]{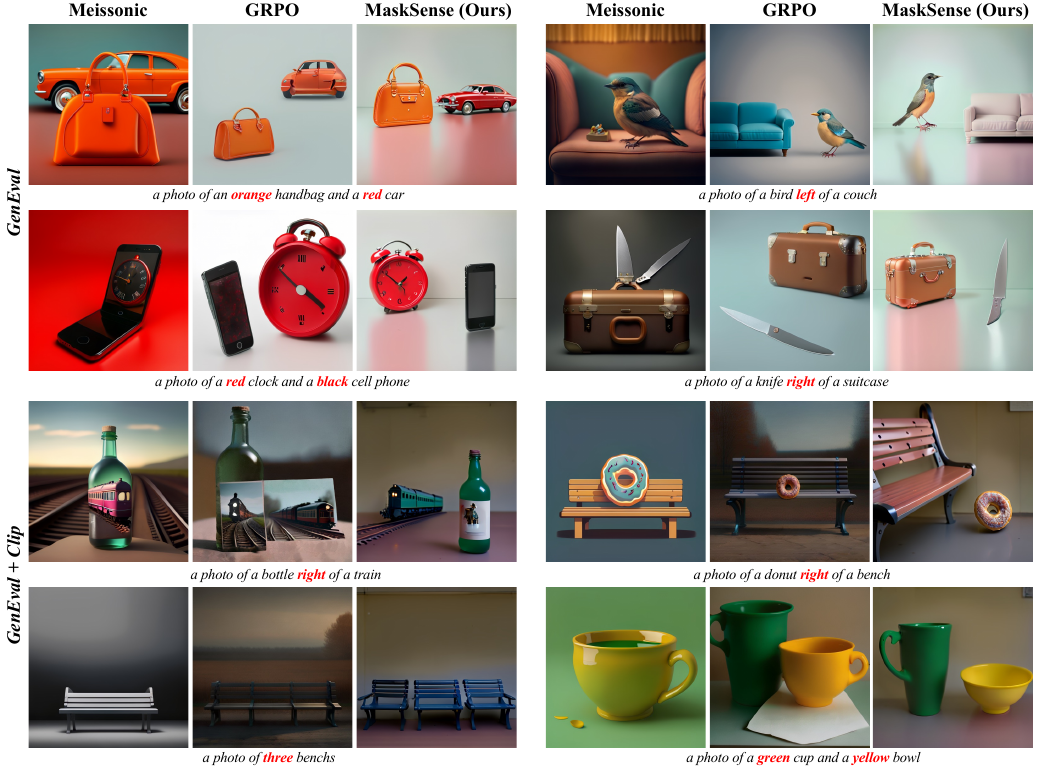}
\end{overpic}
\caption{\textbf{Qualitative Comparisons on Compositional Image Generation.} MaskSense more reliably satisfies compositional constraints including object count, spatial arrangement, and color attribution across GenEval and GenEval+CLIP settings.}
\label{fig:supp_geneval}
\end{figure}

\textbf{More Visualized Results.}
To further validate the effectiveness of MaskSense, we provide additional qualitative comparisons on both the human preference alignment task (Fig.~\ref{fig:supp_hps}) and the compositional image generation task (Fig.~\ref{fig:supp_geneval}), supplementing the visualizations presented in the main paper.

\clearpage
\newpage

\end{document}